\documentclass[10pt,twocolumn,letterpaper]{article}
\usepackage{authblk}
\usepackage[pagenumbers]{iccv} 

\usepackage{tabularx} 
\newcolumntype{Y}{>{\centering\arraybackslash}X}
\newcolumntype{V}{>{\centering\arraybackslash}m{1.9cm}}
\newcolumntype{Z}{>{\centering\arraybackslash}m{1.1cm}}

\usepackage[ruled,vlined]{algorithm2e}
\usepackage{booktabs}
\usepackage{multirow}
\usepackage{subcaption}
\usepackage{indentfirst}
\usepackage{rotating} 
\usepackage{cases}
\usepackage{tablefootnote}
\usepackage{threeparttable}
\usepackage{nicematrix, tikz}
\usepackage{makecell}
\usepackage{stfloats} 
\usepackage{soul}
\newcommand*\mr[2]{\multirow{#1}{*}[-0.5\dimexpr \aboverulesep + \belowrulesep + \cmidrulewidth]{#2}}
\newcommand*\mrn[3]{\multirow{#1}{*}[-#2\dimexpr \aboverulesep + \belowrulesep + \cmidrulewidth]{#3}}
\let\originaleqref\eqref
\renewcommand*{\eqref}[1]{Eq.~\originaleqref{#1}}

\definecolor{iccvblue}{rgb}{0.21,0.49,0.74}
\usepackage[pagebackref,breaklinks,colorlinks,allcolors=iccvblue]{hyperref}

\def\paperID{12468} 
\def\confName{ICCV}
\def\confYear{2025}

\title{EA-KD: Entropy-based Adaptive Knowledge Distillation\vspace{-12pt}}

\author[1,*]{Chi-Ping Su}
\author[2,*]{Ching-Hsun Tseng}
\author[3,\dag]{Bin Pu}
\author[4]{Lei Zhao}
\author[3]{Jiewen Yang}
\author[3]{Zhuangzhuang Chen}
\author[1,\dag]{Shin-Jye Lee}
\affil[ ]{%
    $^{1}$National Yang Ming Chiao Tung University\\
    $^{2}$The University of Manchester\\
}
\affil[ ]{%
    $^{3}$Hong Kong University of Science and Technology\\
    $^{4}$Hunan University\vspace{-18pt}
}

\begin{document}
\maketitle
{
\renewcommand{\thefootnote}{\fnsymbol{footnote}}
\footnotetext[1]{Equal contribution.} 
\footnotetext[2]{Bin Pu and Shin-Jye Lee are corresponding authors (eebinpu@ust.hk, camhero@gmail.com).} 
}
\begin{abstract}
Knowledge distillation (KD) enables a smaller ``student" model to mimic a larger ``teacher" model by transferring knowledge from the teacher's output or features. However, most KD methods treat all samples uniformly, overlooking the varying learning value of each sample and thereby limiting effectiveness. In this paper, we propose \textbf{\underline{E}}ntropy-based \textbf{\underline{A}}daptive \textbf{\underline{K}}nowledge \textbf{\underline{D}}istillation (EA-KD), a simple yet effective plug-and-play KD method that prioritizes learning from valuable samples. EA-KD quantifies each sample’s learning value by strategically combining the entropy of the teacher and student output, then dynamically reweights the distillation loss to place greater emphasis on high-entropy samples. Extensive experiments across diverse KD frameworks and tasks\---including image classification, object detection, and large language model (LLM) distillation\---demonstrate that EA-KD consistently enhances performance, achieving state-of-the-art results with negligible computational cost. Our code is available at \url{https://github.com/cpsu00/EA-KD}.
\vspace{-10pt}
\end{abstract}    
\section{Introduction}
\label{sec:intro}
\vspace{-4pt}
The growing size of state-of-the-art (SOTA) deep learning models poses challenges for deployment in resource-constrained settings. Knowledge Distillation (KD) \cite{kd} offers a solution by training a smaller ``student'' model to mimic a larger ``teacher'' model, using both ground-truth and the teacher’s ``knowledge'' (\eg logits or feature representations) to achieve similar performance in a compact form.
KD has been widely applied across domains, including computer vision (CV) \cite{deit, DINO, dinov2}, large language models (LLMs) \cite{distilbert, tinybert, minillm, distillm}, and medical imaging \cite{p1, p2, p3, p4}.

\begin{figure}[t]
  \centering
   \includegraphics[width=1.0\linewidth]{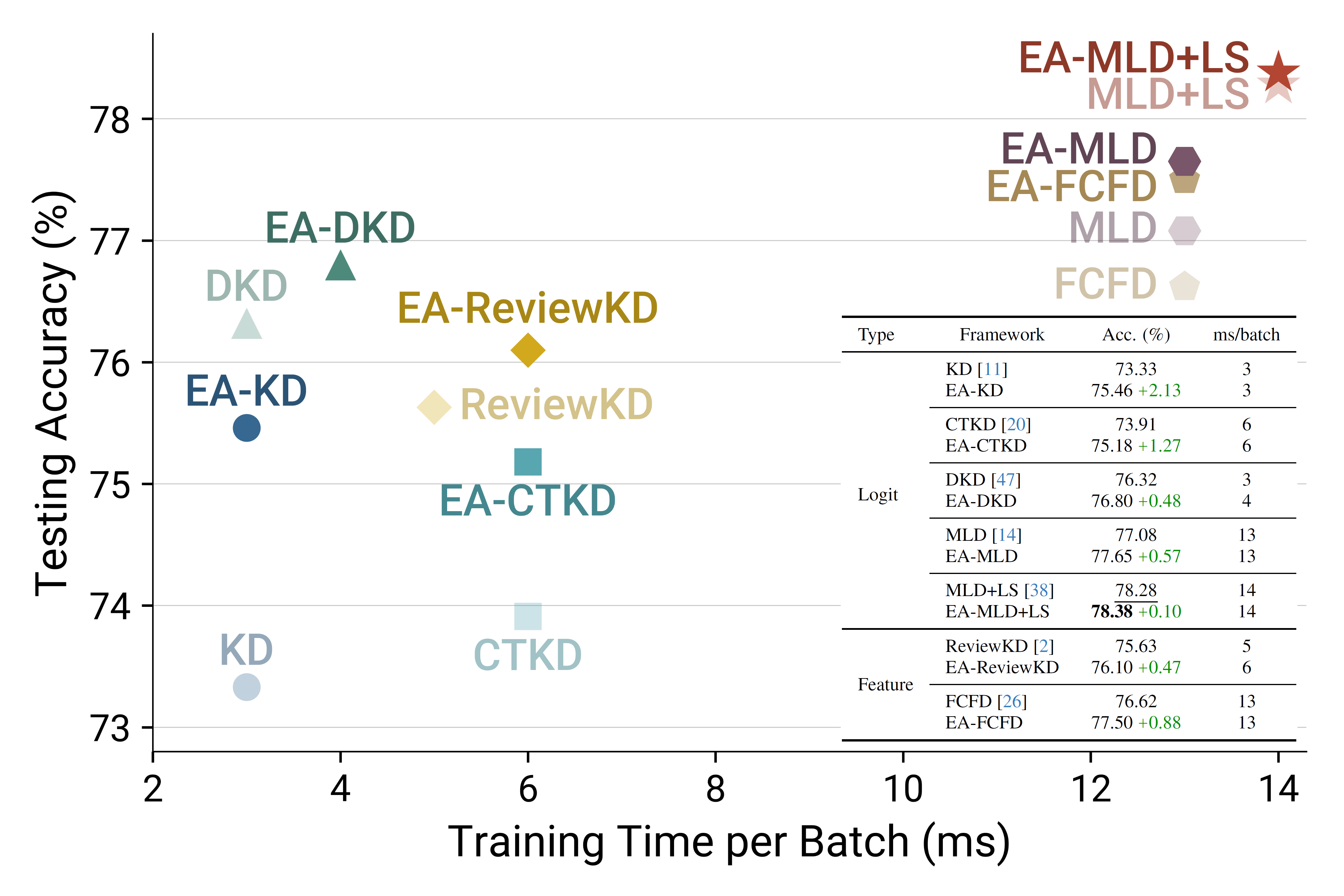}
   \vspace{-18pt}
    \caption{\textbf{Training Time per Batch (ms) \vs Accuracy (\%) of EA-Methods and Baselines on CIFAR-100.}}
   \label{fig:acc_time}
   \vspace{-10pt}
\end{figure}

Advanced KD methods have explored diverse knowledge forms and structural modifications to refine knowledge transfer \cite{kd, fitnet, reviewkd, TAKD, dkd, mld, fcfd}. However, most methods distill uniformly across all samples, operating under the assumption that each sample has equal importance and overlooking their varying learning value. Recent advancements, such as Instance-T \cite{CTKD}, have shown that assigning unique temperatures to each sample outperforms the uniform temperature method Global-T. This highlighted the benefits of adapting to each sample’s distinct characteristics. Building on this, we hypothesize that emphasizing samples rich in \textit{valuable knowledge}\footnote{Throughout this paper, we consider samples with high entropy as containing valuable knowledge.} can further optimize the distillation process. This mirrors how human students learn better when key points are highlighted by teachers. Such a strategy enables the student model to focus on more informative samples, leading to improved performance. However, as we will show, the uniform distillation scheme in most KD methods often overlooks these critical samples, thereby limiting the efficiency of knowledge transfer.

\begin{figure}
\centering
\subcaptionbox{Accuracy \vs Entropy Segments.\label{fig:acc_vs_entropy}}{\includegraphics[height=7.7\baselineskip]{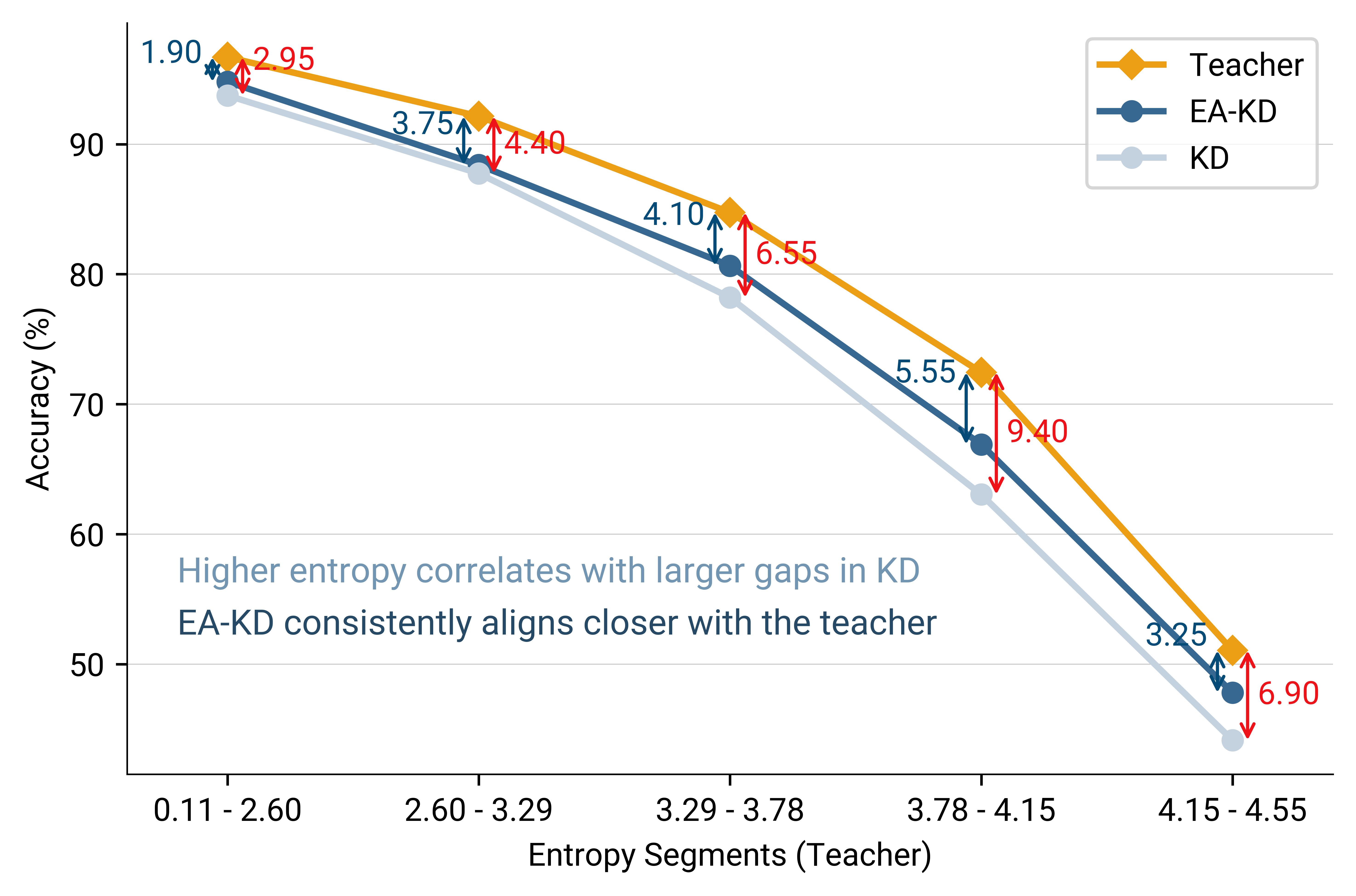}} \quad
\hspace{-10pt}
\subcaptionbox{t-SNE of Teacher.\label{fig:tsne_t}}{\includegraphics[height=7.7\baselineskip]{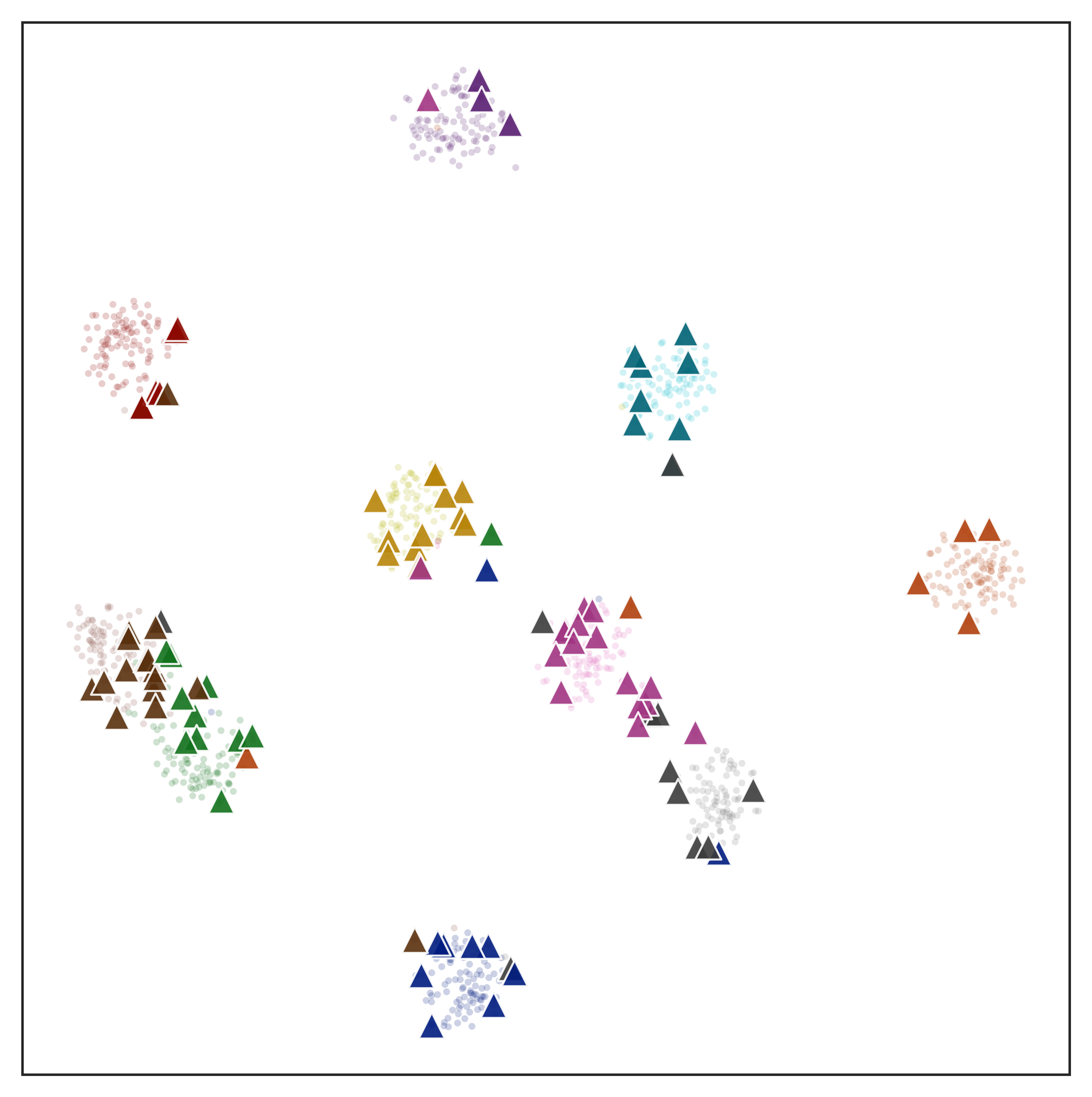}} \quad
\vspace{-4pt}
\caption{
\textbf{High-Entropy Samples in KD.} (a) Higher teacher entropy correlates with larger accuracy gaps in KD, while EA-KD maintains closer alignment. (b) The top 10\% high entropy samples (denoted by triangles) cluster near decision boundaries, representing critical knowledge essential for classification.
}
\label{fig:entropy_in_kd}
\end{figure}

The question then arises: \textit{\textbf{How can we identify the most valuable samples for learning?}} Entropy, the core concept in information theory that quantifies the uncertainty or information of a random variable \cite{entropy}, may be well-suited for this role. 
Prior methods have incorporated entropy in KD to adjust weighting \cite{AKD} or refine logit predictions \cite{dynamickd}; however, these approaches are typically restricted to specific scenarios (\eg, multi-teacher or logit-based KD) and primarily rely on teacher entropy, leaving the broader potential of entropy in KD underexplored.
Therefore, we propose leveraging entropy to quantify the learning value of each sample in KD, as high-entropy outputs shall correspond to greater information contents that are crucial for learning. Our preliminary analysis reveals that higher entropy samples\footnote{Results for high teacher entropy samples are shown here; high student entropy plots are in the \cref{app:high_Hs}.} (i) correlate with larger teacher-student accuracy gaps (\cref{fig:acc_vs_entropy}) and (ii) often lie near class boundaries in t-SNE visualizations \cite{tsne} (\cref{fig:tsne_t}), suggesting these informative samples not only offer valuable learning opportunities but are also pivotal in defining decision boundaries. Thus, entropy can serve as a reliable metric for identifying the most valuable samples in KD.

\begin{table}[t]
\centering
\footnotesize
\caption{
\textbf{Entropy-based Reweighting on KD.} Reweighting with entropy ($H^\mathcal{T}$ and $H^\mathcal{S}$) improves student's performance, while inverted reweighting ($H_{\text{ub}} - w$) reduces accuracy.
}
\begin{tabularx}{0.8\linewidth}{lYY}
\toprule
 & \multicolumn{2}{c}{Reweighting Factor $w$} \\
\cmidrule(l){2-3}
\multicolumn{1}{c}{Loss Function} & $H^\mathcal{T}$ & $H^\mathcal{S}$ \\ 
\midrule
$L_{\text{KD}}$ \cite{kd} & \multicolumn{2}{c}{73.33} \\
$w L_{\text{KD}}$ & \textbf{75.14} {\scriptsize\color[HTML]{008F00}\texttt{+}1.81} & \textbf{74.76} {\scriptsize\color[HTML]{008F00}\texttt{+}1.43} \\
$(H_{\text{ub}} - w) L_{\text{KD}}$ & 72.73 {\scriptsize\color[HTML]{9C0006}\texttt{-}0.60} & 68.90 {\scriptsize\color[HTML]{9C0006}\texttt{-}4.43} \\
\bottomrule
\end{tabularx}
\label{table:reweighting_factors}
\vspace{-6pt}
\end{table}


Leveraging this insight, adapting the focus of KD to valuable samples should fill the need for enhanced distillation. To validate this, we conducted a preliminary experiment that compared the performance of reweighting with teacher ($H^\mathcal{T}$) and student ($H^\mathcal{S}$) entropy, along with their linear-inverted variants ($H_{\text{ub}} - w$), where high-entropy samples received lower weight. Here, \(H_{\text{ub}}\) denotes the upper bound of entropy (\ie, $\log100$ for CIFAR-100). As shown in \cref{table:reweighting_factors}, reweighting with either $H^\mathcal{T}$ or $H^\mathcal{S}$ significantly improved KD accuracy, while inverted reweighting led to decreased performance. This supports our hypothesis that focusing on samples with valuable knowledge enhances student learning, with $H^\mathcal{T}$ proving more effective due to the teacher’s more reliable assessment of sample value.

\begin{figure}[t]
  \centering
  \includegraphics[width=1.0\linewidth]{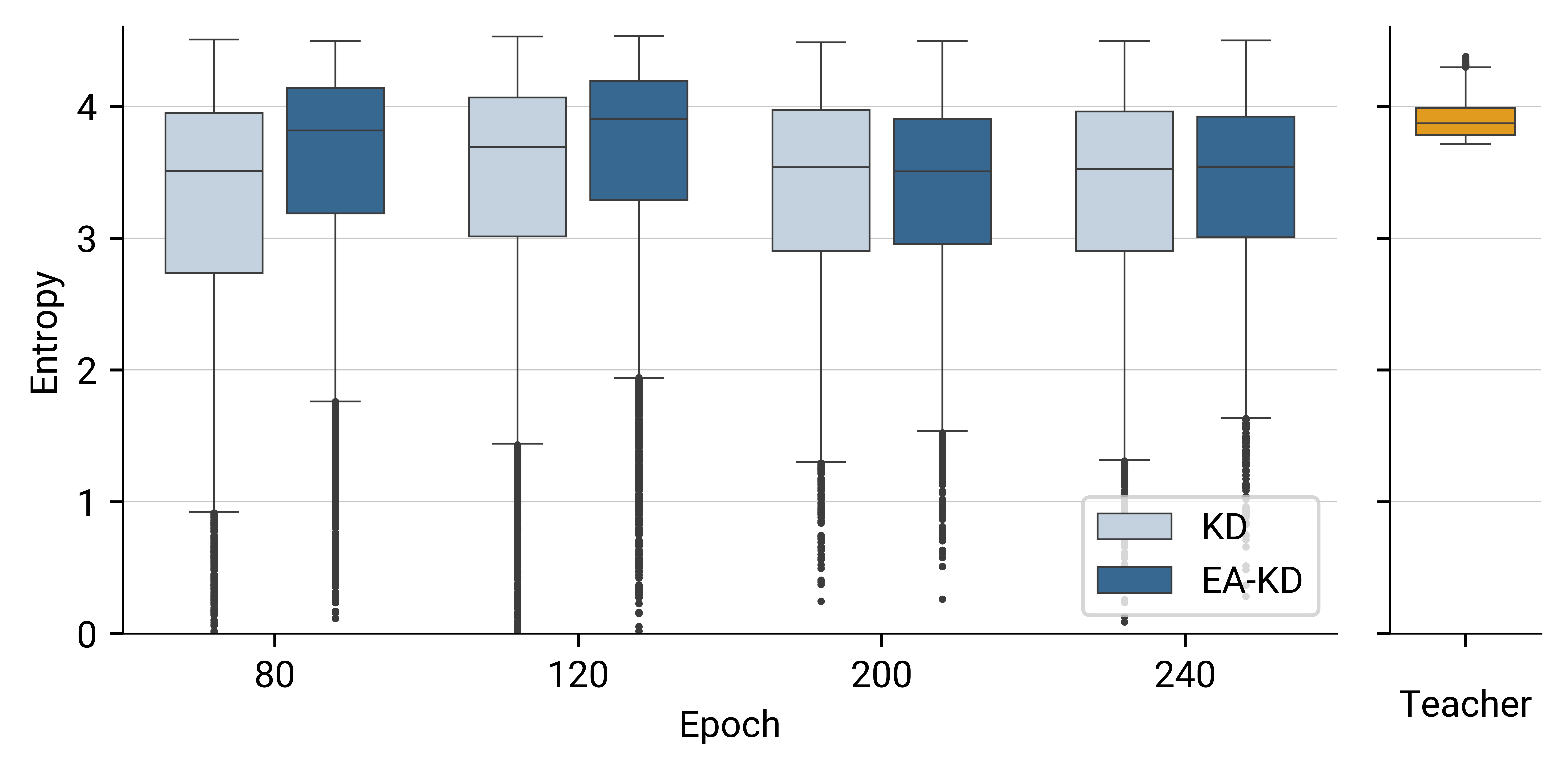}
  \vspace{-16pt} 
  \caption{\textbf{Box Plot of $H^\mathcal{S}$ \vs $H^\mathcal{T}$ for Top 10\% High $H^\mathcal{T}$ Samples.} $H^\mathcal{S}$ shows notable variation, reflecting the student’s learning dynamics and differing perspective from the teacher. EA-KD enhances mimicry with more stable and aligned entropy.}
  \label{fig:boxplot}
   \vspace{-12pt}
\end{figure}

Exploring deeper, we observed that $H^\mathcal{S}$ exhibited increasing variability across training epochs for the top 10\% $H^\mathcal{T}$ samples (\cref{fig:boxplot}). This suggests that while the teacher finds these samples valuable, potentially due to the inherent differences in architecture or capacity, the student’s assessment fluctuates during training. Some samples remain consistently challenging (high $H^\mathcal{S}$), while others become progressively simpler (low $H^\mathcal{S}$) over time. This misalignment reveals the limitation of reweighting solely with $H^\mathcal{T}$, as it remains constant throughout the epochs and thus fails to capture the student’s evolving learning process. 

In light of the above analysis, we introduce Entropy-based Adaptive Knowledge Distillation (EA-KD), a simple yet effective plug-and-play KD method that enhances distillation by reweighting the loss toward high-value samples.
Leveraging both teacher and student entropies, EA-KD emphasizes samples that encapsulate critical knowledge as identified by the teacher, while dynamically adapting to the student’s evolving needs. 
This method directly addresses a fundamental limitation in standard KD frameworks, which often overlook the unique learning value of each sample and tend to bias towards simpler knowledge. 
Furthermore, EA-KD can be seamlessly integrated into most KD frameworks, enhancing their performance with negligible computational cost (\cref{fig:acc_time}). Extensive experiments on image classification, object detection, and LLM distillation demonstrate its efficacy and versatility across diverse KD frameworks. Our main contributions are summarized as follows:
\begin{itemize}
\item We reveal that high-entropy samples carry critical knowledge in KD and propose an entropy-based reweighting factor that integrates both teacher and student entropy to provide a dynamic and tailored learning focus.
\item We introduce EA-KD, a plug-and-play KD method that adaptively reweights the distillation loss to prioritize valuable samples, enabling more effective and efficient knowledge transfer.
\item We demonstrate that EA-KD consistently improves performance across logit- and feature-based KD methods, achieving SOTA results on both CV and LLM tasks with minimal computational overhead.
\end{itemize}

\section{Related Work}
\vspace{-4pt}
\label{sec:relatedwork}
\noindent\textbf{Logit and Feature Distillation.}
Logit distillation \cite{kd, dkd, mld} aligns the softened output logits of the teacher and student, valued for its simplicity and broad applicability. On the other hand, feature distillation minimizes divergence in intermediate feature representations, offering enhanced learning but often with higher computational costs \cite{fitnet,reviewkd,fcfd}. Both pathways have achieved SOTA performance across tasks and domains. However, most of them typically adopt a static distillation scheme, such as treating all samples uniformly. Adaptive distillation addresses this limitation by introducing more dynamic knowledge transfer processes \cite{RWKD, PAD, CTKD, LS, AKD, dynamickd, ttm}

\noindent\textbf{Adaptive Distillation.} 
These methods improve knowledge transfer by dynamically adjusting knowledge at different levels.
For sample-level, RW-KD \cite{RWKD} employs meta-learning to optimize weights for each sample, which introduces high computational overhead. PAD \cite{PAD} showed that traditional hard-mining weighting is unsuitable for KD and instead prioritizes samples with low uncertainty and small teacher-student gaps. This shares similarities with EA-KD, as we will show that high-entropy samples often exhibit lower KLD. However, PAD relies on an auxiliary estimation branch, whereas EA-KD offers a more interpretable and efficient approach by directly utilizing entropy.  
For logit-level, CTKD \cite{CTKD} and LS \cite{LS} dynamically adjust the temperature parameter \(T\) to refine knowledge transfer, but remain limited to logit-based KD. Instead, EA-KD’s sample-wise reweighting ensures broader compatibility across KD frameworks. Importantly, EA-KD and these methods serve distinct yet complementary roles, combining them could further improve performance, as we will show.

\noindent\textbf{Entropy in KD.}
Cheng et al. introduced an entropy-based metric to quantify knowledge retention in KD \cite{quantify}. Inspired by this, we utilize entropy to identify valuable samples in KD. 
AKD \cite{AKD} assign higher weights to low-entropy teacher predictions in multi-teacher settings. However, such weighting can degrade performance in the more common single-teacher settings (\cref{table:reweighting_factors}). 
DynamicKD \cite{dynamickd} refines logit-level knowledge through entropy correction similar to CTKD and LS, and is also constrained to logit-based KD.
TTM \cite{ttm} removes the student temperature, revealing an inherent Rényi entropy regularization, while WTTM further emphasizes the uncertain samples. 
In contrast, EA-KD actively leverages both teacher and student entropy for dynamic weighting, yielding stronger performance\footnote{See \cref{sec:ttm-appendix} for a more detailed comparison.}.

\section{Methodology}
\vspace{-2pt}
\label{sec:method}
\subsection{Preliminaries}
\vspace{-2pt}
\noindent\textbf{Information Theory.}
Entropy quantifies the uncertainty or information content of a random variable \cite{entropy}. For a given sample \(x_n\), the entropy \(H_n\) is computed as follows: 
\begin{equation} \label{eq1}
    H_n = -\sum_{i=1}^{C} \sigma(z_{n,i}) \log(\sigma(z_{n,i}))\,,
\end{equation}
where \(\sigma(\cdot)\) denotes the softmax function, \( z_{n,i} \) is the logit for class \( i \) of sample \( x_n \), and \( C \) is the number of classes.

\vspace{2pt}
\noindent\textbf{Knowledge Distillation.}
The goal of vanilla KD is to transfer the knowledge encapsulated in the teacher’s softened probability outputs to the student \cite{kd}. In classification tasks, the probabilities $p$ are softened using the temperature-scaled softmax function:
\begin{equation} \label{eq2}
    p_i(T) = \sigma(z, T)_i = \frac{\exp(\frac{z_i}{T})}{\sum_{k=1}^{C} \exp(\frac{z_k}{T})}\,,
\end{equation}
where \( p_i(T) \) denotes the softened probability for class \( i \), and \(\sigma(z_i, T)\) is the temperature-scaled softmax function. The temperature $T$ controls the smoothness of the distribution, revealing the subtle inter-class relationships.

The core of KD is to minimize the Kullback-Leibler divergence (KLD) between the teacher’s and student’s softened probabilities, the KD loss is defined as:
\begin{equation} \label{eq3}
\begin{split}
L_{\text{KD}} & = \text{KLD}(p^\mathcal{T}(T) \| p^\mathcal{S}(T)) \cdot T^2\\
              & = \sum_{i=1}^{C} p^\mathcal{T}_i(T) \log\left( \frac{p^\mathcal{T}_i(T)}{p^\mathcal{S}_i(T)} \right) \cdot T^2\,,
\end{split}
\end{equation}
where $p^\mathcal{T}$ and $p^\mathcal{S}$ denote the teacher’s and student’s softened outputs, respectively. For simplicity of theoretical analysis, we set $T = 1$ in this section. $L_{\text{KD}}$ then simplifies to:
\begin{equation} \label{eq4}
\begin{split}
L_{\text{KD}} = \sum_{i=1}^{C} p^\mathcal{T}_i \log\left( \frac{p^\mathcal{T}_i}{p^\mathcal{S}_i} \right).
\end{split}
\end{equation}

\begin{figure*}[t]
    \centering
    \includegraphics[width=.9\linewidth]{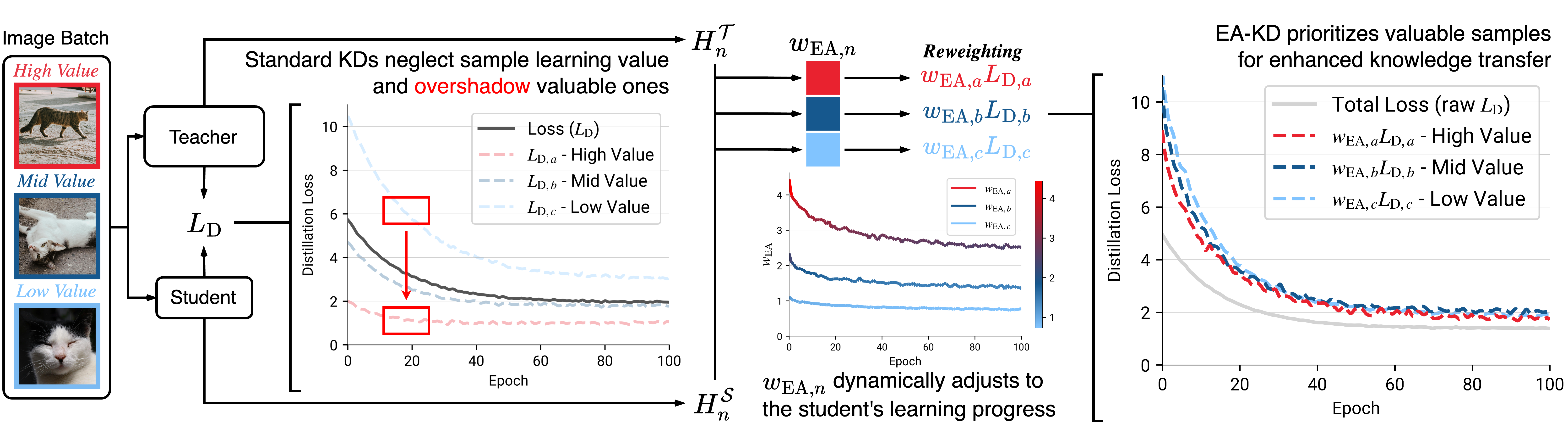}
    \caption{
    \textbf{Illustration of the Uniform Distillation Scheme in Standard KD (Left) and Entropy-based Reweighting in EA-KD (Right).} 
    Standard KD not only overlook the varying learning value of individual samples but also bias toward low-entropy (easier) ones.
    In contrast, EA-KD effectively guides learning toward valuable samples based on both teacher and student assessments.
    }
    \label{fig:illustration}
   \vspace{-12pt}
\end{figure*}

\vspace{-4pt}
\subsection{EA-KD}
\vspace{-2pt}

\noindent\textbf{Limitations in Standard KDs.} Most logit- and feature-based KD methods \cite{kd, dkd, mld, reviewkd, fcfd, fitnet} treat all samples uniformly, overlooking their unique learning value. This oversight can cause the model to over-prioritize simpler samples at the expense of more valuable, high-entropy ones. Taking KLD---the main loss function for logit-based methods---as an example, consider a student initialized with a uniform distribution \( p^\mathcal{S} \) where \( p^\mathcal{S}_i = \frac{1}{C} \, \forall \, i \). For a low-entropy sample with a teacher output \( p^\mathcal{T}_{\text{low}} \) where \( p^\mathcal{T}_{\text{low}, j} \approx 1 \) and \( p^\mathcal{T}_{\text{low}, i} \approx 0 \, (i \neq j) \), the KLD becomes:
\begin{equation}
\begin{split}
    \text{KLD}(p^\mathcal{T}_{\text{low}} \| p^\mathcal{S}) & \approx p^\mathcal{T}_{\text{low}, j} \cdot \log \left( \frac{p^\mathcal{T}_{\text{low}, j}}{p^\mathcal{S}_j} \right) \\
                & = \log \left( \frac{1}{p^\mathcal{S}_j} \right) \\
                & = \log(C).
\end{split}
\end{equation}
For a high-entropy distribution \( p^\mathcal{T}_{\text{high}} \approx \frac{1}{C} \, \forall \, i \), the KLD is:
\begin{equation}
\begin{split}
    \text{KLD}(p^\mathcal{T}_{\text{high}} \| p^\mathcal{S}) & \approx \sum_{i=1}^{C} \frac{1}{C} \cdot \log \left( \frac{1/C}{1/C} \right) \\
                & = 0.
\end{split}
\end{equation}
Thus, we obtain:
\begin{equation}
    \text{KLD}(p^\mathcal{T}_{\text{low}} \| p^\mathcal{S}) > \text{KLD}(p^\mathcal{T}_{\text{high}} \| p^\mathcal{S}).
\end{equation}
This inequality highlights KLD’s inherent bias toward low-entropy samples, which dominate the loss over the valuable, high-entropy ones. 
Similarly, the MSE loss in feature-based KD also biases learning toward high-magnitude activations, which typically correspond to low-entropy samples. Such imbalance shifts the learning focus toward simpler samples, hindering the transfer of knowledge crucial for learning, especially during early training stages (refer to the overshadowing effect in the left part of \cref{fig:illustration}).

\vspace{8pt}
\noindent\textbf{Entropy-based Reweighting.} As discussed in \cref{sec:intro}, entropy can serve as a measure of the sample learning value in KD. In practice, we soften the entropy with an alternative temperature $T'$ to better reflect sample value:
\begin{equation} \label{eq_H}
H_n = -\sum_{i=1}^{C} p_{n,i}(T') \log(p_{n,i}(T'))\,,
\end{equation}
where $p_{n,i}(T')$ is the temperature-scaled probability of class $i$ for sample $x_n$. 

To dynamically emphasize valuable samples, EA-KD’s reweighting factor $w_{\text{EA}}$ is formulated using two components: a base term $w_{\text{base}}$ and an interaction term $w_{\text{interact}}$. The base term captures the inherent value of each sample based on the teacher’s output entropy, defined as:
\begin{equation} \label{eq_teacher_term}
w_{\text{base},n} = H^\mathcal{T}_n, \quad H^\mathcal{T}_n \in [0, H_{\text{ub}}]\,,
\end{equation}
where \( H^\mathcal{T}_n \) denotes the entropy of the teacher’s prediction for sample \( x_n \), and \( H_{\text{ub}} = \log(C) \) is the upper bound of entropy for \( C \) classes. The interaction term $w_{\text{interact}}$, on the other hand, captures the interplay between the teacher and student perspectives by taking the normalized product of their entropies:
\begin{equation} \label{eq_interaction_term}
w_{\text{interact},n} = \frac{H^\mathcal{T}_n \cdot H^\mathcal{S}_n}{H_{\text{ub}}}, \quad w_{\text{interact},n} \in [0, H_{\text{ub}}]\,.
\end{equation}
where \( H^\mathcal{S}_n \) is the student’s entropy. Finally, the EA-KD reweighting factor $w_{\text{EA},n}$ is defined as the average of the base and interaction terms:
\begin{equation} \label{eq_wn_avg}
w_{\text{EA},n} = \frac{w_{\text{base},n} + w_{\text{interact},n}}{2}, \quad w_{\text{EA},n} \in [0, H_{\text{ub}}]\,.
\end{equation}
By integrating both the teacher’s evaluation and the student’s evolving understanding, $w_{\text{EA}}$ effectively prioritizes valuable samples while dynamically adjusting the focus throughout training.

\vspace{4pt}
\noindent\textbf{Reformulation.} 
To better illustrate the influence of the student’s perspective in $w_{\text{EA},n}$, we reformulate \cref{eq_wn_avg} as:
\begin{equation} \label{eq_reformulated_wn}
\begin{split}
    w_{\text{EA},n} & = \frac{H^\mathcal{T}_n + \frac{H^\mathcal{T}_n \cdot H^\mathcal{S}_n}{H_{\text{ub}}}}{2} \\
    & = \frac{1}{2} H^\mathcal{T}_n \left(1 + \frac{H^\mathcal{S}_n}{H_{\text{ub}}}\right)\,.
\end{split}
\end{equation}
This reformulation expresses \( w_{\text{EA},n} \) as the product of $H^\mathcal{T}_n$ and a scaling factor that depends on the $H^\mathcal{S}_n$. Thus, \( w_{\text{EA},n} \) can be regarded as the teacher’s assessment of sample value, adaptively adjusted based on the student’s perspective. Depending on the entropy values, $w_{\text{EA}}$ behaves as follows:
\begin{subnumcases}{w_{\text{EA},n} = }
    H_{\text{ub}} & if $H^\mathcal{T}_n \rightarrow H_{\text{ub}} \land H^\mathcal{S}_n \rightarrow H_{\text{ub}}$ \label{eq:c1} \\
    \frac{H_{\text{ub}}}{2} & if $H^\mathcal{T}_n \rightarrow H_{\text{ub}} \land H^\mathcal{S}_n \rightarrow 0$ \label{eq:c2} \\
    0 & if $H^\mathcal{T}_n \rightarrow 0 \quad \forall H^\mathcal{S}_n$ \label{eq:c3} \\
    w_{\text{EA},n} & otherwise. \notag
\end{subnumcases}
When the teacher considers a sample highly valuable for learning (\( H^\mathcal{T}_n \rightarrow H_{\text{ub}} \)), the scaling factor adjusts \( w_{\text{EA},n} \) based on the student’s view. If the student aligns (\eqref{eq:c1}), the weight is maximized to emphasize this sample. Conversely, if the student is confident (\eqref{eq:c2}), the weight reduces to half for moderate focus. However, when the teacher considers the sample simple (\eqref{eq:c3}), the weight remains low regardless of \( H^\mathcal{S}_n \), as the teacher considers it lacks valuable knowledge.

\begin{table*}[ht]
\centering
\footnotesize  
\setlength{\tabcolsep}{5pt} 
\begin{threeparttable}
\caption{\textbf{Results on CIFAR-100.} Accuracy (\%) of EA-methods \vs baselines across teacher-student pairs, with relative improvements highlighted. Avg. $\Delta$ shows average improvement across methods and model pairs. Best results are \textbf{bolded}, and second-best are \underline{underlined}.}
\label{tab:cifar100}

\begin{tabularx}{\textwidth}{clYYYcYYYc@{}}
\toprule
\multicolumn{1}{c}{\multirow{4}{*}{Type}} & \multicolumn{1}{c}{\multirow{2}{*}{Teacher}} & ResNet32$\times$4 & WRN-28-4 & WRN-40-2 & VGG13 & VGG13 & ResNet50 & ResNet32$\times$4 &  \\
& & 79.42 & 78.60 & 75.61 & 74.64 & 74.64 & 79.34 & 79.42 &  \\
& \multicolumn{1}{c}{\multirow{2}{*}{Student}} & ResNet8$\times$4 & WRN-16-2 & WRN-40-1 & VGG8 & MN-V2 & MN-V2 & SN-V2 &  \\
& & 72.50 & 73.26 & 71.98 & 70.36 & 64.60 & 64.60 & 71.82 & Avg. \(\Delta\) \\ 
\midrule

\mrn{8}{4}{Logit} & KD \cite{kd} & 73.33 & 75.04 & 73.54 & 72.98 & 67.37 & 67.35 & 74.45 & \\
& EA-KD & 75.46 {\scriptsize\color[HTML]{008F00}\texttt{+}2.13} & 75.79 {\scriptsize\color[HTML]{008F00}\texttt{+}0.75} & 74.38 {\scriptsize\color[HTML]{008F00}\texttt{+}0.84} & 74.08 {\scriptsize\color[HTML]{008F00}\texttt{+}1.10} & 69.17 {\scriptsize\color[HTML]{008F00}\texttt{+}1.80} & 69.67 {\scriptsize\color[HTML]{008F00}\texttt{+}2.32} & 75.91 {\scriptsize\color[HTML]{008F00}\texttt{+}1.46} & {\color[HTML]{008F00}\texttt{+}1.48\phantom{+}} \\

\cmidrule{2-10}

& CTKD \cite{CTKD} & 73.91 & 75.29 & 73.93 & 73.52 & 68.46 & 68.47 & 75.31 & \\ 
& EA-CTKD & 75.18 {\scriptsize\color[HTML]{008F00}\texttt{+}1.27} & 75.72 {\scriptsize\color[HTML]{008F00}\texttt{+}0.43} & 74.03 {\scriptsize\color[HTML]{008F00}\texttt{+}0.10} & 73.79 {\scriptsize\color[HTML]{008F00}\texttt{+}0.27} & 69.19 {\scriptsize\color[HTML]{008F00}\texttt{+}0.73} & 69.38 {\scriptsize\color[HTML]{008F00}\texttt{+}0.91} & 76.02 {\scriptsize\color[HTML]{008F00}\texttt{+}0.71} & {\color[HTML]{008F00}\texttt{+}0.63\phantom{+}} \\

\cmidrule{2-10}

& DKD \cite{dkd} & 76.32 & 76.45 & 74.81 & 74.68 & 69.71 & 70.35 & 77.07 & \\
& EA-DKD & 76.80 {\scriptsize\color[HTML]{008F00}\texttt{+}0.48} & 76.74 {\scriptsize\color[HTML]{008F00}\texttt{+}0.29} & 74.98 {\scriptsize\color[HTML]{008F00}\texttt{+}0.17} & 75.07 {\scriptsize\color[HTML]{008F00}\texttt{+}0.39} & 70.39 {\scriptsize\color[HTML]{008F00}\texttt{+}0.68} & 70.98 {\scriptsize\color[HTML]{008F00}\texttt{+}0.63} & 77.72 {\scriptsize\color[HTML]{008F00}\texttt{+}0.65} & {\color[HTML]{008F00}\texttt{+}0.47\phantom{+}} \\

\cmidrule{2-10}

& MLD \cite{mld} & 77.08 & 76.83 & 75.35 & 75.18 & 70.57 & 71.04 & 78.44 &  \\ 
& EA-MLD & 77.65 {\scriptsize\color[HTML]{008F00}\texttt{+}0.57} & \underline{77.47} {\scriptsize\color[HTML]{008F00}\texttt{+}0.64} & \underline{75.77} {\scriptsize\color[HTML]{008F00}\texttt{+}0.42} & 75.28 {\scriptsize\color[HTML]{008F00}\texttt{+}0.10} & 70.72 {\scriptsize\color[HTML]{008F00}\texttt{+}0.15} & \underline{71.43} {\scriptsize\color[HTML]{008F00}\texttt{+}0.39} & \underline{78.85} {\scriptsize\color[HTML]{008F00}\texttt{+}0.41} & {\color[HTML]{008F00}\texttt{+}0.38\phantom{+}} \\

\cmidrule{2-10}

& MLD+LS \cite{LS} & \underline{78.28} & 77.20 & 75.56 & 75.22 & \underline{70.94} & 71.19 & 78.76 &  \\
& EA-MLD+LS & \textbf{78.38} {\scriptsize\color[HTML]{008F00}\texttt{+}0.10} & \textbf{77.60} {\scriptsize\color[HTML]{008F00}\texttt{+}0.39} & \textbf{75.78} {\scriptsize\color[HTML]{008F00}\texttt{+}0.22} & \textbf{75.38} {\scriptsize\color[HTML]{008F00}\texttt{+}0.16} & 70.67 {\scriptsize\color[HTML]{9C0006}\texttt{-}0.27} & 71.36 {\scriptsize\color[HTML]{008F00}\texttt{+}0.17} & \textbf{79.13} {\scriptsize\color[HTML]{008F00}\texttt{+}0.37} & {\color[HTML]{008F00}\texttt{+}0.16\phantom{+}} \\

\midrule

\mr{4}{Feature} & ReviewKD \cite{reviewkd} & 75.63 & 76.39 & 74.45 & 74.45 & 70.37 & 69.89 & 77.78 &  \\
& EA-ReviewKD & 76.10 {\scriptsize\color[HTML]{008F00}\texttt{+}0.47} & 76.95 {\scriptsize\color[HTML]{008F00}\texttt{+}0.56} & 75.43 {\scriptsize\color[HTML]{008F00}\texttt{+}0.98} & 74.56 {\scriptsize\color[HTML]{008F00}\texttt{+}0.11} & 70.55 {\scriptsize\color[HTML]{008F00}\texttt{+}0.18} & 69.80 {\scriptsize\color[HTML]{9C0006}\texttt{-}0.09} & 78.22 {\scriptsize\color[HTML]{008F00}\texttt{+}0.44} & {\color[HTML]{008F00}\texttt{+}0.38\phantom{+}} \\

\cmidrule{2-10}

& FCFD \cite{fcfd} & 76.62 & 77.00 & 75.46 & 75.22 & 70.65 & 71.00 & 78.18 & \\ 
& EA-FCFD & 77.50 {\scriptsize\color[HTML]{008F00}\texttt{+}0.88} & 77.15 {\scriptsize\color[HTML]{008F00}\texttt{+}0.15} & 75.30 {\scriptsize\color[HTML]{9C0006}\texttt{-}0.16} & \underline{75.36} {\scriptsize\color[HTML]{008F00}\texttt{+}0.14} & \textbf{71.02} {\scriptsize\color[HTML]{008F00}\texttt{+}0.37} & \textbf{71.97} {\scriptsize\color[HTML]{008F00}\texttt{+}0.97} & 78.75 {\scriptsize\color[HTML]{008F00}\texttt{+}0.56} & {\color[HTML]{008F00}\texttt{+}0.42\phantom{+}} \\

\midrule
& Avg. \(\Delta\) & {\color[HTML]{008F00}\texttt{+}0.84} & {\color[HTML]{008F00}\texttt{+}0.46} & {\color[HTML]{008F00}\texttt{+}0.37} & {\color[HTML]{008F00}\texttt{+}0.33} & {\color[HTML]{008F00}\texttt{+}0.52} & {\color[HTML]{008F00}\texttt{+}0.76} & {\color[HTML]{008F00}\texttt{+}0.66} & {\color[HTML]{008F00}\texttt{+}0.56\phantom{+}} \\ \bottomrule
\end{tabularx}
\end{threeparttable}
\vspace{-10pt}
\end{table*} 








\noindent\textbf{Loss Integration.} 
As shown in \cref{fig:illustration}, the flexible nature of $w_{\text{EA}}$ allows it to be plug-and-play into most distillation frameworks by reweighting the contribution of each sample based on its learning value. For instance, when integrated with vanilla KD, the loss is defined as:
\begin{equation} \label{eq_EA}
\begin{split}
    L_{\text{EA-KD}} & = \sum_{n=1}^{N} w_{\text{EA},n}  \cdot L_{\text{KD},n} ,
\end{split}
\end{equation}
where \( L_{\text{KD},n} \) is the distillation loss for sample \( x_n \). As a result, EA-KD enhances standard KD methods by facilitating a more nuanced and adaptive transfer of knowledge, ensuring that informative samples receive increased focus throughout training.


\section{Experiments}
\label{sec:experiments}
\vspace{-2pt}
\subsection{Experimental Setup}
\vspace{-2pt}
\noindent\textbf{Datasets.} 
We used CIFAR-100 \cite{cifar} (50k training, 10k validation images; 100 classes), Tiny-ImageNet \cite{tiny-imagenet} (100k training, 5k validation images; 200 classes), and ImageNet \cite{imagenet} (1.28M training, 50k validation images; 1,000 classes) to evaluate our method for image classification. For object detection, we adopted MS-COCO \cite{coco} (118k training, 5k validation images; 80 classes). Furthermore, for LLM distillation, we employed five instruction-following datasets following \cite{minillm}: Dolly \cite{dolly}, SInst \cite{self-inst}, Vicuna \cite{vicuna}, S-NI \cite{s-ni}, and UnNI \cite{u-ni}.

\begin{table*}[t]
\footnotesize  
\centering
\caption{\textbf{Results on Tiny-ImageNet.} Accuracy (\%) of the ResNet32\(\times\)4 teacher and ResNet8\(\times\)4 student.}
\vspace{-4pt}
\begin{tabularx}{1\textwidth}{cccYcYcYcY}
\toprule
Teacher & Student & KD & \makebox[0pt][c]{EA-KD} & MLD & \makebox[0pt][c]{EA-MLD} & MLD+LS & \makebox[0pt][c]{EA-MLD+LS} & FCFD & \makebox[0pt][c]{EA-FCFD} \\
\midrule
64.41 & 55.25 & 56.00 & 59.39 {\scriptsize\color[HTML]{008F00}\texttt{+}3.39} & 61.91 & \textbf{62.65} {\scriptsize\color[HTML]{008F00}\texttt{+}0.74} & 61.36 & \underline{62.41} {\scriptsize\color[HTML]{008F00}\texttt{+}1.05} & 60.12 & 60.51 {\scriptsize\color[HTML]{008F00}\texttt{+}0.39}  \\
\bottomrule
\end{tabularx}
\label{tab:tintable}
\vspace{-4pt}
\end{table*}

\begin{table*}[t]
\footnotesize  
\centering
\caption{\textbf{Results on ImageNet.} Accuracy (\%) of the ResNet34 teacher and ResNet18 student, averaged over three runs.}
\vspace{-4pt}
\begin{tabularx}{1\textwidth}{ccYYYYYcccY}
\toprule
Teacher & Student & KD \cite{kd} & EA-KD & KD+LS \cite{LS} & DKD \cite{dkd} & EA-DKD & DKD+LS \cite{LS} & EA-DKD+LS & PAD \cite{PAD} \\
\cmidrule(){1-2} \cmidrule(lr){3-5} \cmidrule(lr){6-9} \cmidrule(){10-10}
73.31 & 69.75 & 71.03 & \textbf{71.79} {\scriptsize\color[HTML]{008F00}\texttt{+}0.76} & \ul{71.42} & 71.70 & \ul{71.96} {\scriptsize\color[HTML]{008F00}\texttt{+}0.26} & 71.88 & \textbf{71.99} {\scriptsize\color[HTML]{008F00}\texttt{+}0.11} & 71.71 \\
\bottomrule
\end{tabularx}
\label{tab:imagenet}
\vspace{-8pt}
\end{table*}

\begin{table}[t]
\footnotesize
\centering
\caption{\textbf{Results on Tiny-ImageNet with Transformers.} Accuracy (\%) of transformer-based teachers and a ResNet8\(\times\)4 student, averaged over three runs.}
\vspace{-4pt}
\begin{tabularx}{\linewidth}{YYcYc}
\toprule
\multicolumn{2}{c}{Teacher} & Student & KD & EA-KD \\
\midrule
ViT-B  & 71.12 & \multirow{3}{*}{\makecell{ResNet8$\times$4\\ 55.25}} & 54.70 & \textbf{57.53} {\scriptsize\color[HTML]{008F00}\texttt{+}2.83} \\
DeiT-B & 85.55 & & 56.30 & \textbf{58.36} {\scriptsize\color[HTML]{008F00}\texttt{+}2.06} \\
Swin-B & 86.30 & & 56.15 & \textbf{59.58} {\scriptsize\color[HTML]{008F00}\texttt{+}3.43} \\
\bottomrule
\end{tabularx}
\label{tab:tintable2}
\end{table}

\begin{table}[t]
\footnotesize
\centering
\caption{\textbf{Results on MS-COCO.} AP (overall), AP\(_{50}\), and AP\(_{75}\) are reported using Faster R-CNN with FPN.}
\vspace{-4pt}
\setlength{\tabcolsep}{5pt} 
\begin{tabularx}{\linewidth}{cYYYYYYY}
\toprule
& \multicolumn{3}{c}{R-101 \& R-18} & \multicolumn{3}{c}{R-50 \& MV2} \\
\cmidrule(r){2-4} \cmidrule(l){5-7}
& AP & AP\(_{50}\) & AP\(_{75}\) & AP & AP\(_{50}\) & AP\(_{75}\) \\
\midrule[\heavyrulewidth]
Teacher & 42.04 & 62.48 & 45.88 & 40.22 & 61.02 & 43.81 \\
Student & 33.26 & 53.61 & 35.26 & 29.47 & 48.87 & 30.90 \\ 
\midrule
KD & 33.97 & 54.66 & 36.62 & 30.13 & 50.28 & 31.35 \\
EA-KD & \textbf{34.78} & \textbf{56.14} & \textbf{37.19} & \textbf{31.81} & \textbf{53.18} & \textbf{33.18} \\
$\Delta$ & \color[HTML]{008F00}\texttt{+}0.81 & \color[HTML]{008F00}\texttt{+}1.48 & \color[HTML]{008F00}\texttt{+}0.57 & \color[HTML]{008F00}\texttt{+}1.68 & \color[HTML]{008F00}\texttt{+}2.90 & \color[HTML]{008F00}\texttt{+}1.83 \\
\bottomrule
\end{tabularx}
\label{tab:cocotable}
\vspace{-8pt}
\end{table}

\vspace{2pt}
\noindent\textbf{KD Frameworks.} We evaluated EA-KD across representative SOTA logit-based (KD \cite{kd}, CTKD \cite{CTKD}, DKD \cite{dkd}, MLD \cite{mld}, MLD+LS \cite{LS}) and feature-based methods (ReviewKD \cite{reviewkd}, FCFD \cite{fcfd}), reweighting their distillation loss while preserving each framework’s original structure. For fair comparisons, no hyperparameter was tuned for the EA-variants, except for the KD weight of MLD+LS \cite{LS} was reduced from 9.0 to 4.0 to avoid over-penalizing.

\vspace{2pt}
\noindent\textbf{Implementation Details.} EA-KD and other variants were evaluated on various CNNs (VGG \cite{vgg}, ResNet \cite{resnet}, WideResNet \cite{wideresnet}, MobileNet \cite{mobilenet}, ShuffleNet \cite{shufflenet}), transformer teachers (ViT \cite{vit}, DeiT \cite{deit}, Swin \cite{swin}), and GPT2 \cite{gpt2} for LLM distillation. We used the training settings of \cite{crd} for vanilla KD and the respective settings of each framework for classification, \cite{reviewkd} for detection, and \cite{dskd} for LLMs. We set $T'=3$ based on our ablation study, except $T'=2$ for ImageNet and transformer teachers. All results averaged over five runs, and analyses were conducted using ResNet32\(\times\)4-ResNet8\(\times\)4 pair on CIFAR-100.
\vspace{-4pt}

\subsection{Results}
\vspace{-4pt}
\noindent\textbf{CIFAR-100.} \cref{tab:cifar100} presents results of various EA-methods and their baselines on CIFAR-100. EA-KD consistently improves both logit- and feature-based KD frameworks across most teacher-student pairs, with an average gain of 0.56\%. The logit-based EA-MLD+LS achieved SOTA results in most pairings, while EA-FCFD excelled in two heterogeneous pairs. Additionally, EA-MLD and EA-FCFD secured most second-best results, outperforming the previous SOTA, MLD+LS \cite{LS}. These findings highlight EA-KD’s broad applicability and effectiveness in enhancing knowledge transfer across diverse types of KD frameworks.

\vspace{4pt}
\noindent\textbf{Tiny-ImageNet and ImageNet.} \cref{tab:tintable} and \cref{tab:imagenet} show that EA-methods consistently outperform their baselines on both Tiny-Imagenet and ImageNet. On Tiny-ImageNet, EA-KD improves KD by 3.39\% and achieves performance comparable to the novel FCFD \cite{fcfd}. On ImageNet, EA-KD surpasses other adaptive KDs, including logit-level adaptive LS \cite{LS} and sample-level adaptive approach PAD \cite{PAD}.
These results highlight EA-methods' scalability on larger, more diverse datasets. Furthermore, \cref{tab:tintable2} underscores EA-KD’s advantage in distilling valuable knowledge from vision transformer-based teachers to CNN students. Unlike standard KD, which plateaued when using DeiT-B as the teacher, EA-KD consistently improved student performance, mitigating the capacity gap and utilizing the guidance of stronger teachers more effectively.

\vspace{4pt}
\noindent\textbf{MS-COCO.} Extending to object detection, \cref{tab:cocotable} presents the performance of EA-KD and KD on the MS-COCO dataset. EA-KD consistently outperforms KD across AP metrics for both model pairings, with particularly notable gains in AP$_{50}$. This highlights EA-KD’s ability to effectively guide the student with valuable knowledge in a more complex visual task, thereby enhancing its performance.

\begin{table}[t]
\centering
\footnotesize
\setlength{\tabcolsep}{5pt}
\caption{\textbf{LLM Distillation Results.} Rouge-L \cite{rL} scores averaged over five seeds on each dataset are reported. SFT denotes a student supervised fine-tuned on the dataset.}
\vspace{-4pt}
\label{tab:llm}
\begin{tabularx}{\linewidth}{cYYYYYY}
\toprule
& \multicolumn{5}{c}{Dataset} \\
\cmidrule(lr){2-6}
Method & Dolly & SInst & Vicuna & S-NI & UnNI & Avg. \\ 
\midrule
\makecell{Teacher\\GPT2-XL} & 27.19 & 14.64 & 16.30 & 27.55 & 31.42 & 23.42 \\ 
\makecell{SFT\\GPT2-S} & 22.94 & 10.11 & 15.17 & 16.21 & 18.68 & 16.62 \\ 
\midrule
KD & \underline{24.54} & 10.43 & 15.66 & 17.24 & 20.28 & 17.63 \\ 
RKLD & 24.38 & \textbf{10.73} & \underline{15.71} & \underline{17.31} & \underline{20.96} & \underline{17.82} \\
JSD & 23.86 & 10.20 & 15.50 & 16.20 & 19.17 & 16.98 \\
\midrule
EA-KD & \textbf{24.95} & \underline{10.59} & \textbf{16.41} & \textbf{18.27} & \textbf{21.46} & \textbf{18.34} \\
$\Delta$ to KD & {\color[HTML]{008F00}\texttt{+}0.41} & {\color[HTML]{008F00}\texttt{+}0.16} & {\color[HTML]{008F00}\texttt{+}0.75} & {\color[HTML]{008F00}\texttt{+}1.03} & {\color[HTML]{008F00}\texttt{+}1.18} & {\color[HTML]{008F00}\texttt{+}0.71} \\
\bottomrule
\end{tabularx}
\vspace{-8pt}
\end{table}

\vspace{4pt}
\noindent\textbf{LLM Distillation.}
To explore the full potential of EA-KD, we extended our experiments to LLM distillation, applying the reweighting at the sequence level of the model outputs. As shown in \cref{tab:llm}, EA-KD consistently outperforms standard KD across datasets. Moreover, EA-KD also surpasses two recently emerging methods in this field, Reverse KLD (RKLD) and Jensen-Shannon divergence (JSD), which aim to address KLD’s limitation of forcing the student to cover all modes of the complex LLM teacher distribution \cite{minillm}. In this scenario, EA-KD’s dynamic prioritization of the most critical knowledge performed effectively in such complexity. This highlights EA-KD's versatility beyond visual tasks, showcasing its ability to target valuable knowledge across diverse domains.

\subsection{Empirical Analysis}
\label{sec:analysis}
In this section, we conduct ablation studies to evaluate two key components of EA-KD: $T'$ and the reweighting factor. We then compare EA-KD and vanilla KD from multiple perspectives, showing how EA-KD improves teacher-student alignment and class separability. Additionally, we discuss the synergy between EA and DKD, demonstrating enhanced robustness in both the loss surface and hyperparameter stability. Finally, we highlight EA-methods computational efficiency across diverse KD frameworks.
\vspace{2pt}


\noindent\textbf{Ablation Study.} Two experiment were performed to evaluate the sensitivity of $T'$ and the effect of each reweighting components in $w_{\text{EA}}$. (i) \cref{tab:tp_table} shows that a $T'$ of 3 consistently delivers optimal performance across model combinations, highlighting its robustness in reflecting sample value. (ii) As shown in \cref{table:reweighting_factors2}, $w_{\text{base}}$ enhances KD by emphasizing valuable samples under the teacher’s guidance. While $w_{\text{interact}}$ integrates student dynamics, it may introduce noise and dilute the teacher’s guidance, reducing effectiveness. However, when integrated into $w_\text{EA}$, $H^\mathcal{S}$ acts as a scaling factor for $H^\mathcal{T}$ (\cref{eq_reformulated_wn}). This ensures early training, where $H^\mathcal{S}$ is mostly high and the scaling near 1, is mainly guided by $H^\mathcal{T}$. As training progresses, $H^\mathcal{S}$ adaptively adjusts and tailors the weighting, leading to superior performance.

\begin{table}[t]
\centering
\footnotesize
\caption{\textbf{Impact of \(T'\).} Setting \(T'\) to 3 yields the best results across architectures, demonstrating its robustness and consistency.}
\vspace{-4pt}
\begin{tabularx}{\linewidth}{ccYYY}
\toprule
\mr{2}{Teacher} & \mr{2}{Student} & \multicolumn{3}{c}{$T'$} \\ 
\cmidrule(lr){3-5} 
 & & 2 & 3 & 4 \\ \midrule
ResNet32$\times$4 & ResNet8$\times$4 & 73.90 & \textbf{75.46} & \underline{75.22} \\
WRN-28-4 & WRN-16-2 & 74.89 & \textbf{75.79} & \underline{75.62} \\
ResNet32$\times$4 & SN-V2 & 75.51 & \textbf{75.91} & \underline{75.72} \\
\bottomrule
\end{tabularx}
\label{tab:tp_table}
\end{table}

\begin{table}[t]
\centering
\footnotesize
\caption{\textbf{Impact of Reweighting Factors.} The proposed $w_{\text{EA}}$ outperforms the single-sided $w_{\text{base}}$, while $w_{\text{interact}}$, lacking teacher base guidance, shows limited improvement.}
\vspace{-4pt}
\begin{tabularx}{\linewidth}{lYYYY}
\toprule
\multicolumn{1}{c}{\mr{2}{Method}} & \mr{2}{Acc.} & \multicolumn{3}{c}{Reweighting Factor} \\
\cmidrule(lr){3-5}
& & $w_{\text{base}}$ & $w_{\text{interact}}$ & $w_{\text{EA}}$ \\ 
\midrule
KD \cite{kd} & 73.33 & \phantom{$\uparrow$} \underline{75.14} {\scriptsize\color[HTML]{008f00}$\uparrow$} & \phantom{$\uparrow$} 74.76 {\scriptsize\color[HTML]{008f00}$\uparrow$} & \phantom{$\uparrow$} \textbf{75.46} {\scriptsize\color[HTML]{008f00}$\uparrow$} \\ 
MLD \cite{mld} & 77.08 & \phantom{$\uparrow$} \underline{77.47} {\scriptsize\color[HTML]{008f00}$\uparrow$} & \phantom{$\uparrow$} 77.45 {\scriptsize\color[HTML]{008f00}$\uparrow$} & \phantom{$\uparrow$} \textbf{77.65} {\scriptsize\color[HTML]{008f00}$\uparrow$} \\ 
MLD+LS \cite{LS} & 78.28 & \phantom{$\uparrow$} \underline{78.30} {\scriptsize\color[HTML]{008f00}$\uparrow$} & \phantom{$\uparrow$} 78.20 {\scriptsize\color[HTML]{9C0006}$\downarrow$} & \phantom{$\uparrow$} \textbf{78.38} {\scriptsize\color[HTML]{008f00}$\uparrow$} \\ 
FCFD \cite{fcfd} & 76.62 & \phantom{$\uparrow$} 77.50 {\scriptsize\color[HTML]{008f00}$\uparrow$} & \phantom{$\uparrow$} \underline{77.42} {\scriptsize\color[HTML]{008f00}$\uparrow$} & \phantom{$\uparrow$} \textbf{77.44} {\scriptsize\color[HTML]{008f00}$\uparrow$} \\ 
\bottomrule
\end{tabularx}
\label{table:reweighting_factors2}
\vspace{-8pt}
\end{table}

\begin{figure}[t]
    \centering
    \includegraphics[width=1.0\linewidth]{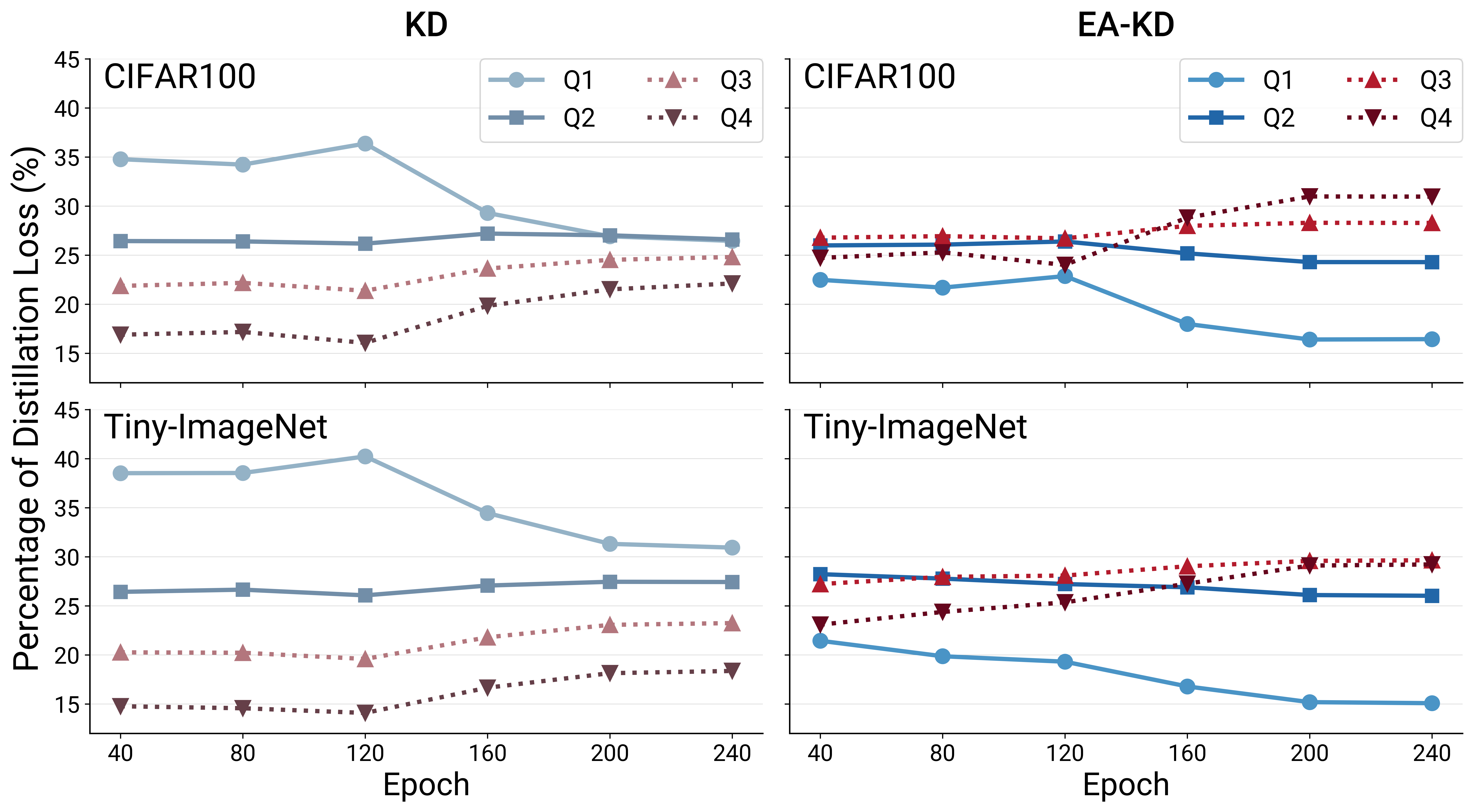}
    \vspace{-18pt}
    \caption{\textbf{Loss Distribution \vs Entropy Quartiles over Epochs.} Low-value samples (Q1, Q2; blue) dominate the KD loss, whereas EA-KD places more focus on high-value ones (Q3, Q4; red).}
    \label{fig:LvsH}
\end{figure}

\vspace{4pt}
\noindent\textbf{Distillation Loss and Sample Value.}
In \cref{fig:LvsH}, we analyze the distribution of distillation loss across samples grouped by low- to high-entropy quartiles (Q1–Q4) for KD and EA-KD on CIFAR-100 and Tiny-ImageNet. 
Notably, KD loss is dominated by low-value samples throughout training on both datasets. The overshadowing effect in KD’s uniform distillation scheme hinders the transfer of critical knowledge from high-value samples, as discussed in \cref{sec:method}, ultimately leading to more significant accuracy gaps (\cref{fig:acc_vs_entropy}).
In contrast, EA-KD emphasizes the focus on valuable samples (\cref{fig:LvsH}), bringing holistic performance improvements across entropy segments (\cref{fig:acc_vs_entropy}).

\begin{figure}[t]
\centering
\includegraphics[width=1\linewidth]{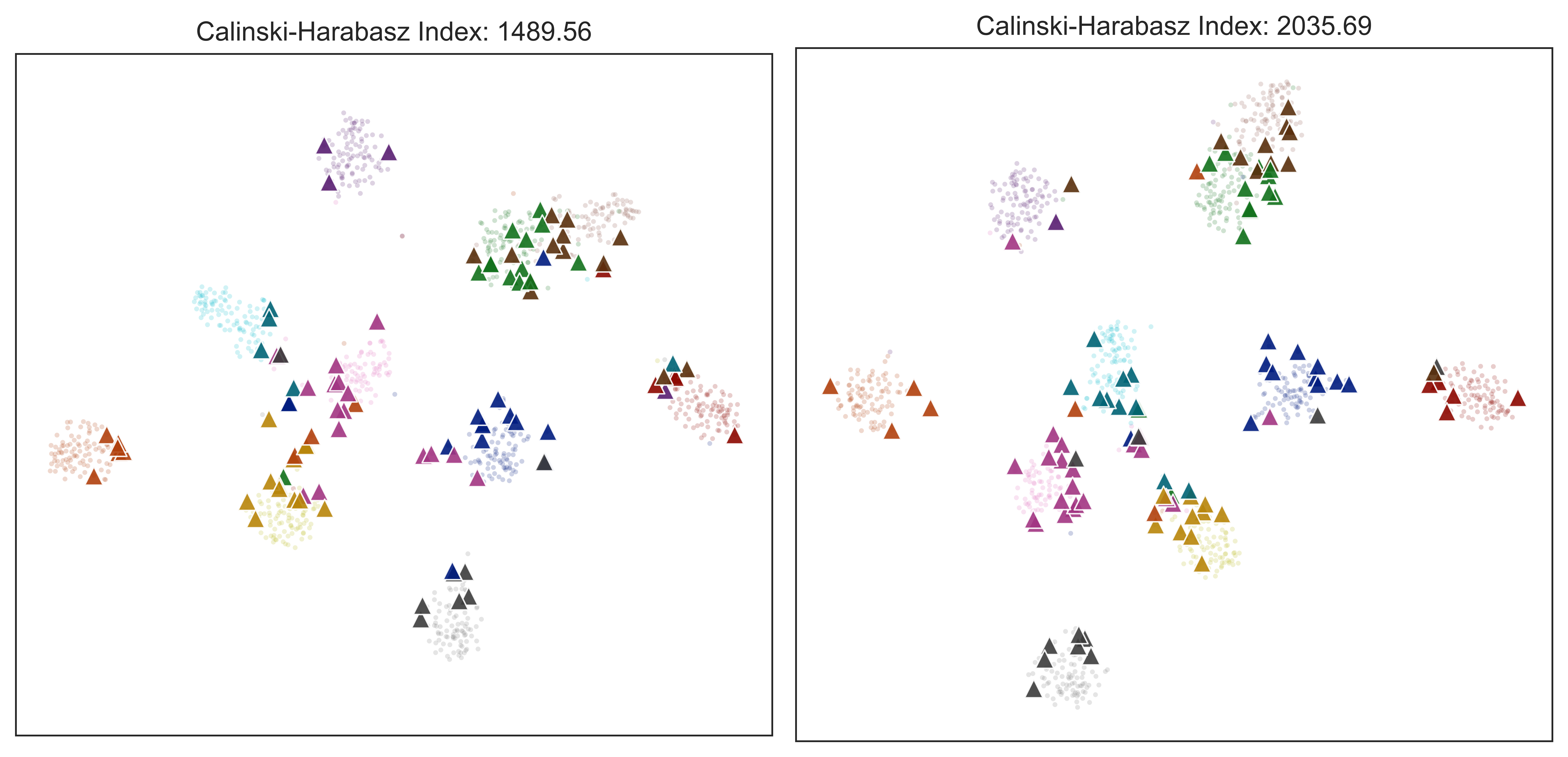}
\caption{
\textbf{t-SNE of Students.} 
Top 10\% high $H^\mathcal{S}$ samples for KD (left) and $w_{\text{EA}}$ samples for EA-KD (right) are highlighted with triangles. High $H^\mathcal{S}$ samples cluster near decision boundaries, underscoring their learning value for the student.
}
\label{fig:tsne}
\vspace{-14pt}
\end{figure}

\vspace{4pt}
\noindent\textbf{Student's Unique Perspective.} \cref{fig:tsne} compares the t-SNE visualizations \cite{tsne} of KD and EA-KD with the high $H^\mathcal{S}$ and $w_{\text{EA}}$ samples highlighted\footnote{See \cref{app:high_Hs} for further t-SNE visualizations showing the student’s evolving sample focus over training epochs.}. Similar to \cref{fig:tsne_t}, high $H^\mathcal{S}$ samples also lie near decision boundaries in the KD-student. This suggests that, despite having a different view from the teacher (\cref{fig:boxplot}), the student’s evolving capacity enables $H^\mathcal{S}$ to capture the samples crucial for its own learning progress. In EA-KD, integrating both $H^\mathcal{T}$ and $H^\mathcal{S}$ in $w_{\text{EA}}$ captures both the teacher’s informed assessment and the student’s dynamic learning needs. As a result, EA-KD achieves superior class separability as reflected by a higher Calinski-Harabasz (CH) index.

\vspace{4pt}
\noindent\textbf{Consistency in Entropy Levels.} As shown in \cref{fig:boxplot}, EA-KD mitigates the increasing entropy variation across epochs observed in KD, enabling the student to maintain more stable $H^\mathcal{S}$ and align more closely to $H^\mathcal{T}$. This stability creates a feedback loop in EA-KD, where a steady $H^\mathcal{S}$ leads to a stable $w_{\text{EA}}$, promoting a more focused learning process. As a result, the student more effectively mimics the teacher's response and achieves improved performance.

\begin{figure*}
    \centering
    \includegraphics[width=1\textwidth]{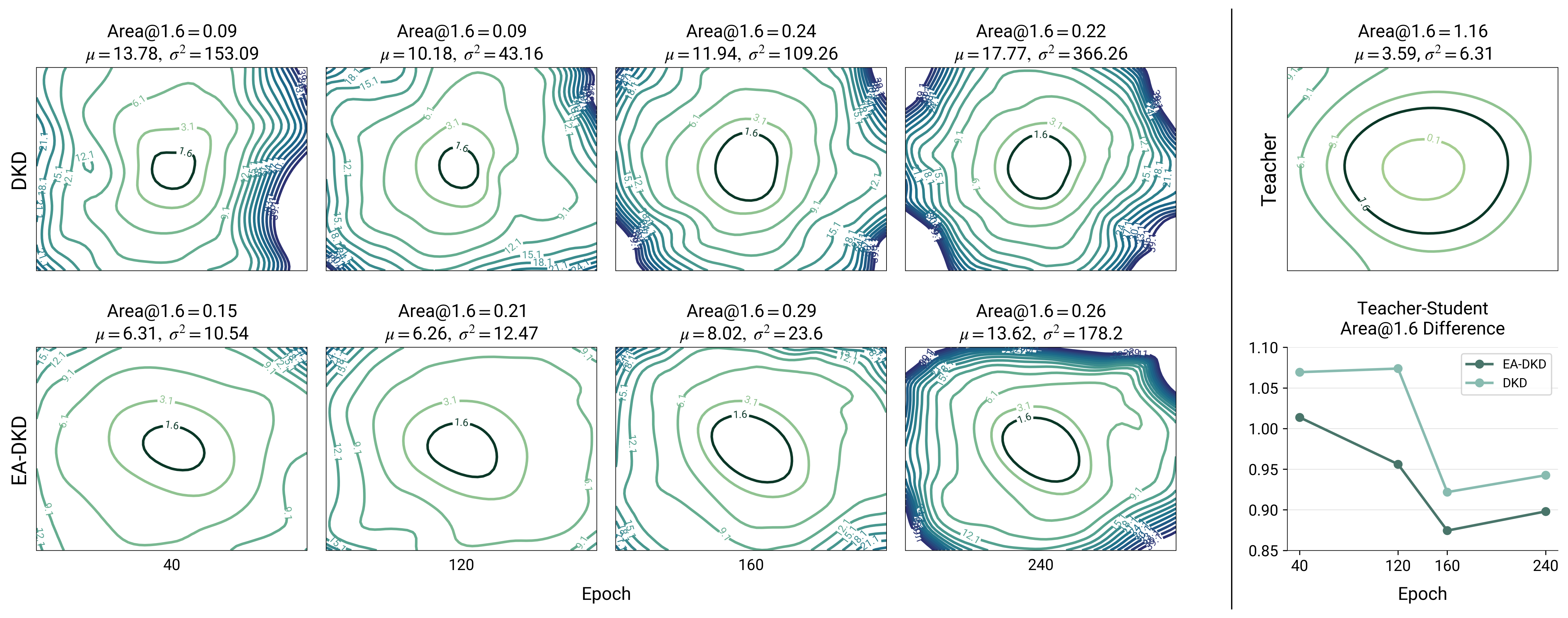}
    \vspace{-20pt}
    \caption{\textbf{Loss Surface and Differences in Area@1.6 for DKD and EA-DKD Students Across Epochs.} The mean and variance for the surface, along with the contour areas at level 1.6 (Area@1.6), are provided for each subplot. The line plot (lower right) tracks the differences in Area@1.6 between the teacher and students over epochs. EA-DKD consistently shows larger contours with less fluctuation, signifying smoother learning surfaces and a more robust generalization process compared to DKD.
    }
    \label{fig:dkd-landscapes}
    \vspace{-8pt}
\end{figure*}

\begin{figure}
    \centering
   \includegraphics[width=1\linewidth]{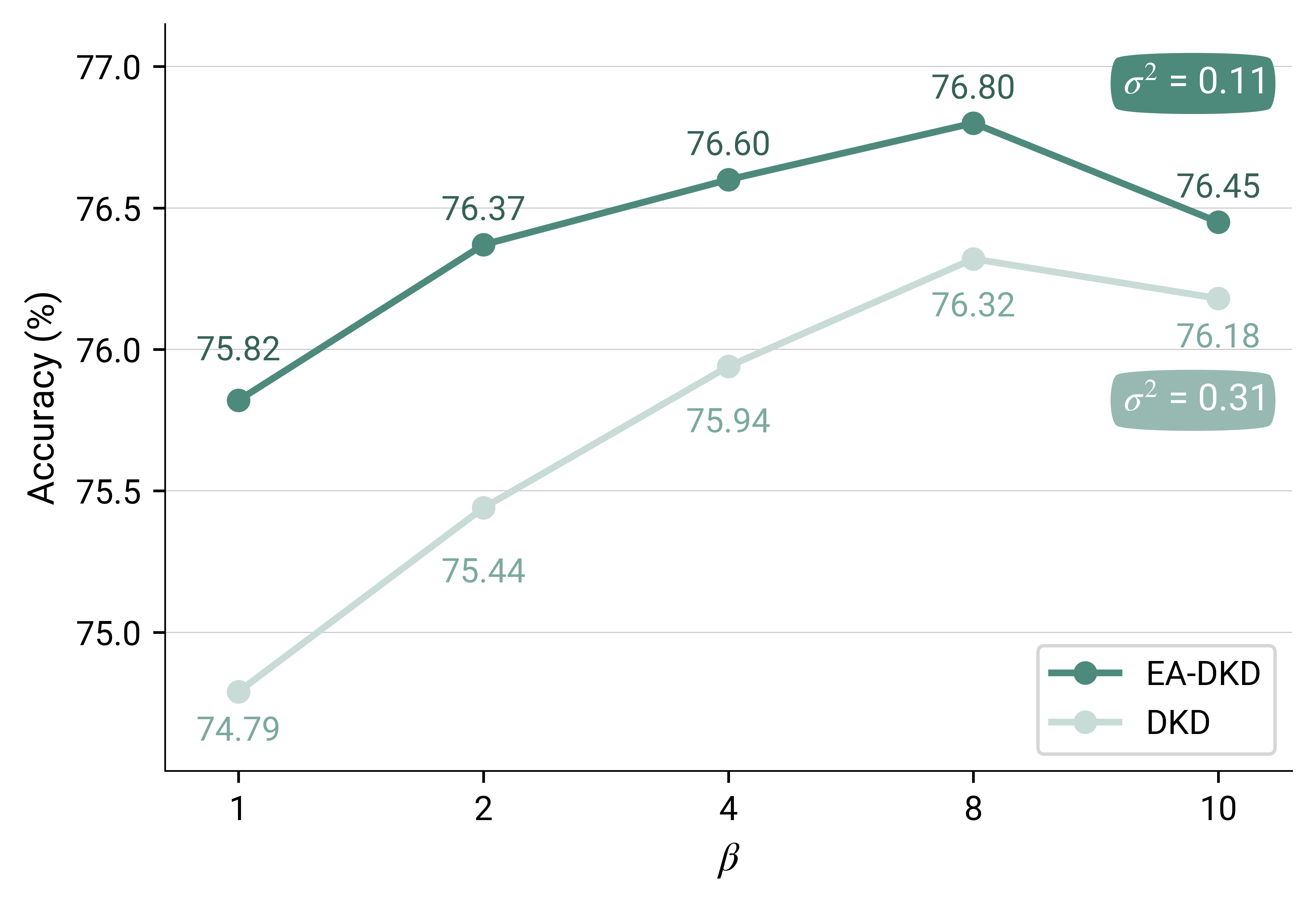}
    \vspace{-20pt}
    \caption{\textbf{Comparison of DKD and EA-DKD Performance Across Varying $\beta$ Values.} EA-DKD consistently outperforms DKD and significantly reduces performance variance over $\beta$ (0.10 \vs 0.31), highlighting EA-DKD's enhanced robustness.}
    \label{fig:dkd-b}
    \vspace{-14pt}
\end{figure}

\vspace{4pt}
\noindent\textbf{Loss Landscape and Synergy in EA-DKD.} Visualizing the loss landscape \cite{visualloss} offers insights into the student's robustness against noise and generalization. As illustrated in \cref{fig:dkd-landscapes}, DKD shows narrow and fluctuating contours with significantly higher variance across epochs (\eg 366.26 at epoch 240), indicating instability in the loss surface and weaker generalization. In contrast, EA-DKD consistently produces a smoother surface with more stable contours (\eg variance of 178.20 at epoch 240) and broader low-loss regions (Area@1.6) compared to DKD, suggesting improved generalizability. In addition, the Area@1.6 comparison (\cref{fig:dkd-landscapes}, lower right) underscores EA-DKD’s closer alignment with the teacher’s surface throughout training.

The enhanced generalization in EA-DKD also extends to hyperparameter robustness. DKD was introduced to decouple target class KD (TCKD) and non-target class KD (NCKD) in vanilla KD, balancing them with hyperparameters \(\alpha\) and \(\beta\). While $\alpha$ remains stable around 1.0, adjusting $\beta$ from 1.0 to 10.0 leads to fluctuating performance \cite{dkd}, with a variance of 0.31 (see \cref{fig:dkd-b}). Notably, EA-DKD effectively reduces this fluctuation to a variance of 0.10 and consistently improves the performance. This enhanced robustness may be attributed to the synergy between DKD and EA in handling different aspects of knowledge transfer: DKD prevents NCKD from being overshadowed by TCKD at the class level, while EA ensures valuable samples---often rich in NCKD---are not dominated by simpler ones at the sample level. Additionally, the dynamic focus in EA compensates for the static nature of $\beta$. Together, EA-DKD facilitates a nuanced and balanced knowledge transfer process, both class-wise and sample-wise.

\vspace{4pt}
\noindent\textbf{Computational Efficiency.}
\cref{fig:acc_time} illustrates the efficiency of EA-methods, 
showing significant performance improvements across various SOTA KD frameworks with negligible cost. This remarkable efficiency, combined with its streamlined integration into both logit- and feature-based KD methods, underscores the potential of our method as a versatile and practical enhancement for KD.
\vspace{-4pt}
\section{Conclusion}
\vspace{-4pt}
In this paper, we revisited existing KD methods from a novel perspective and revealed a key limitation in their inherent uniform distillation strategy, which often hinders the transfer of high-entropy samples that carry critical knowledge. To address this, we proposed EA-KD, a plug-and-play KD approach that dynamically reweights the distillation loss, directing the learning focus toward valuable samples. EA-KD consistently enhances representative KD baselines across image classification, object detection, and LLM distillation, all with negligible cost. We believe EA-KD showcases a great paradigm of the meticulous handling of knowledge transfer, adapting KD to the varying learning value of samples while accounting for the student’s evolving learning dynamics throughout the distillation process.

{
    \small
    \bibliographystyle{ieeenat_fullname}
    \bibliography{main}

\begin{thebibliography}{49}
\providecommand{\natexlab}[1]{#1}
\providecommand{\url}[1]{\texttt{#1}}
\expandafter\ifx\csname urlstyle\endcsname\relax
  \providecommand{\doi}[1]{doi: #1}\else
  \providecommand{\doi}{doi: \begingroup \urlstyle{rm}\Url}\fi

\bibitem[Caron et~al.(2021)Caron, Touvron, Misra, J{\'e}gou, Mairal, Bojanowski, and Joulin]{DINO}
Mathilde Caron, Hugo Touvron, Ishan Misra, Herv{\'e} J{\'e}gou, Julien Mairal, Piotr Bojanowski, and Armand Joulin.
\newblock Emerging properties in self-supervised vision transformers.
\newblock In \emph{Proceedings of the IEEE/CVF international conference on computer vision}, pages 9650--9660, 2021.

\bibitem[Chen et~al.(2021)Chen, Liu, Zhao, and Jia]{reviewkd}
Pengguang Chen, Shu Liu, Hengshuang Zhao, and Jiaya Jia.
\newblock Distilling knowledge via knowledge review.
\newblock In \emph{Proceedings of the IEEE/CVF Conference on Computer Vision and Pattern Recognition}, pages 5008--5017, 2021.

\bibitem[Cheng et~al.(2020)Cheng, Rao, Chen, and Zhang]{quantify}
Xu Cheng, Zhefan Rao, Yilan Chen, and Quanshi Zhang.
\newblock Explaining knowledge distillation by quantifying the knowledge.
\newblock \emph{CoRR}, abs/2003.03622, 2020.

\bibitem[Chiang et~al.(2023)Chiang, Li, Lin, Sheng, Wu, Zhang, Zheng, Zhuang, Zhuang, Gonzalez, Stoica, and Xing]{vicuna}
Wei-Lin Chiang, Zhuohan Li, Zi Lin, Ying Sheng, Zhanghao Wu, Hao Zhang, Lianmin Zheng, Siyuan Zhuang, Yonghao Zhuang, Joseph~E. Gonzalez, Ion Stoica, and Eric~P. Xing.
\newblock Vicuna: An open-source chatbot impressing gpt-4 with 90\%* chatgpt quality, 2023.

\bibitem[Conover et~al.(2023)Conover, Hayes, Mathur, Xie, Wan, Shah, Ghodsi, Wendell, Zaharia, and Xin]{dolly}
Mike Conover, Matt Hayes, Ankit Mathur, Jianwei Xie, Jun Wan, Sam Shah, Ali Ghodsi, Patrick Wendell, Matei Zaharia, and Reynold Xin.
\newblock Free dolly: Introducing the world's first truly open instruction-tuned llm, 2023.

\bibitem[Deng et~al.(2009)Deng, Dong, Socher, Li, Li, and Fei-Fei]{imagenet}
Jia Deng, Wei Dong, Richard Socher, Li-Jia Li, Kai Li, and Li Fei-Fei.
\newblock Imagenet: A large-scale hierarchical image database.
\newblock In \emph{2009 IEEE Conference on Computer Vision and Pattern Recognition}, pages 248--255, 2009.

\bibitem[Dosovitskiy et~al.(2021)Dosovitskiy, Beyer, Kolesnikov, Weissenborn, Zhai, Unterthiner, Dehghani, Minderer, Heigold, Gelly, Uszkoreit, and Houlsby]{vit}
Alexey Dosovitskiy, Lucas Beyer, Alexander Kolesnikov, Dirk Weissenborn, Xiaohua Zhai, Thomas Unterthiner, Mostafa Dehghani, Matthias Minderer, Georg Heigold, Sylvain Gelly, Jakob Uszkoreit, and Neil Houlsby.
\newblock An image is worth 16x16 words: Transformers for image recognition at scale.
\newblock In \emph{International Conference on Learning Representations}, 2021.

\bibitem[Gao et~al.(2025)Gao, Lin, Wen, Pu, Feng, and Li]{p4}
Zhan Gao, Qika Lin, Huaxuan Wen, Bin Pu, Mengling Feng, and Kenli Li.
\newblock Incorporating large vision model distillation and fuzzy perception for improving disease diagnosis.
\newblock \emph{IEEE Transactions on Fuzzy Systems}, 2025.

\bibitem[Gu et~al.(2024)Gu, Dong, Wei, and Huang]{minillm}
Yuxian Gu, Li Dong, Furu Wei, and Minlie Huang.
\newblock Minillm: Knowledge distillation of large language models.
\newblock In \emph{The Twelfth International Conference on Learning Representations}, 2024.

\bibitem[He et~al.(2016)He, Zhang, Ren, and Sun]{resnet}
Kaiming He, Xiangyu Zhang, Shaoqing Ren, and Jian Sun.
\newblock Deep residual learning for image recognition.
\newblock In \emph{Proceedings of the IEEE conference on computer vision and pattern recognition}, pages 770--778, 2016.

\bibitem[Hinton et~al.(2015)Hinton, Vinyals, and Dean]{kd}
Geoffrey Hinton, Oriol Vinyals, and Jeff Dean.
\newblock Distilling the knowledge in a neural network.
\newblock \emph{arXiv preprint arXiv:1503.02531}, 2015.

\bibitem[Honovich et~al.(2022)Honovich, Scialom, Levy, and Schick]{u-ni}
Or Honovich, Thomas Scialom, Omer Levy, and Timo Schick.
\newblock Unnatural instructions: Tuning language models with (almost) no human labor.
\newblock \emph{arXiv preprint arXiv:2212.09689}, 2022.

\bibitem[Jiao et~al.(2019)Jiao, Yin, Shang, Jiang, Chen, Li, Wang, and Liu]{tinybert}
Xiaoqi Jiao, Yichun Yin, Lifeng Shang, Xin Jiang, Xiao Chen, Linlin Li, Fang Wang, and Qun Liu.
\newblock Tinybert: Distilling bert for natural language understanding.
\newblock \emph{arXiv preprint arXiv:1909.10351}, 2019.

\bibitem[Jin et~al.(2023)Jin, Wang, and Lin]{mld}
Ying Jin, Jiaqi Wang, and Dahua Lin.
\newblock Multi-level logit distillation.
\newblock In \emph{Proceedings of the IEEE/CVF Conference on Computer Vision and Pattern Recognition}, pages 24276--24285, 2023.

\bibitem[Ko et~al.(2024)Ko, Kim, Chen, and Yun]{distillm}
Jongwoo Ko, Sungnyun Kim, Tianyi Chen, and Se-Young Yun.
\newblock Distillm: towards streamlined distillation for large language models.
\newblock In \emph{Proceedings of the 41st International Conference on Machine Learning}, pages 24872--24895, 2024.

\bibitem[Krizhevsky(2009)]{cifar}
Alex Krizhevsky.
\newblock Learning multiple layers of features from tiny images.
\newblock Technical report, University of Toronto, 2009.

\bibitem[Kwon et~al.(2020)Kwon, Na, Lee, and Kim]{AKD}
Kisoo Kwon, Hwidong Na, Hoshik Lee, and Nam~Soo Kim.
\newblock Adaptive knowledge distillation based on entropy.
\newblock In \emph{ICASSP 2020 - 2020 IEEE International Conference on Acoustics, Speech and Signal Processing (ICASSP)}, pages 7409--7413, 2020.

\bibitem[Le and Yang(2015)]{tiny-imagenet}
Ya Le and Xuan Yang.
\newblock Tiny imagenet visual recognition challenge.
\newblock \emph{CS 231N}, 7\penalty0 (7):\penalty0 3, 2015.

\bibitem[Li et~al.(2018)Li, Xu, Taylor, Studer, and Goldstein]{visualloss}
Hao Li, Zheng Xu, Gavin Taylor, Christoph Studer, and Tom Goldstein.
\newblock Visualizing the loss landscape of neural nets.
\newblock In \emph{Neural Information Processing Systems}, 2018.

\bibitem[Li et~al.(2023)Li, Li, Yang, Zhao, Song, Luo, Li, and Yang]{CTKD}
Zheng Li, Xiang Li, Lingfeng Yang, Borui Zhao, Renjie Song, Lei Luo, Jun Li, and Jian Yang.
\newblock Curriculum temperature for knowledge distillation.
\newblock In \emph{Proceedings of the AAAI Conference on Artificial Intelligence}, pages 1504--1512, 2023.

\bibitem[Liang et~al.(2025{\natexlab{a}})Liang, Huang, Pu, Chen, Hua, Zhang, Ma, Chen, Li, and Chang]{p1}
Pengchen Liang, Haishan Huang, Bin Pu, Jianguo Chen, Xiang Hua, Jing Zhang, Weibo Ma, Zhuangzhuang Chen, Yiwei Li, and Qing Chang.
\newblock Task-specific knowledge distillation from the vision foundation model for enhanced medical image segmentation.
\newblock \emph{arXiv preprint arXiv:2503.06976}, 2025{\natexlab{a}}.

\bibitem[Liang et~al.(2025{\natexlab{b}})Liang, Shi, Yao, Pu, Chen, Zhao, Huang, Chen, Xu, Xu, et~al.]{p2}
Pengchen Liang, Leijun Shi, Huiping Yao, Bin Pu, Jianguo Chen, Lei Zhao, Haishan Huang, Zhuangzhuang Chen, Zhaozhao Xu, Lite Xu, et~al.
\newblock Rapid bone scintigraphy enhancement via semantic prior distillation from segment anything model.
\newblock \emph{arXiv preprint arXiv:2503.02321}, 2025{\natexlab{b}}.

\bibitem[Liang et~al.(2025{\natexlab{c}})Liang, Shi, Yao, Pu, Chen, Zhao, Huang, Chen, Xu, Xu, et~al.]{p3}
Pengchen Liang, Leijun Shi, Huiping Yao, Bin Pu, Jianguo Chen, Lei Zhao, Haishan Huang, Zhuangzhuang Chen, Zhaozhao Xu, Lite Xu, et~al.
\newblock Semantic prior distillation with vision foundation model for enhanced rapid bone scintigraphy image restoration.
\newblock \emph{arXiv e-prints}, pages arXiv--2503, 2025{\natexlab{c}}.

\bibitem[Lin(2004)]{rL}
Chin-Yew Lin.
\newblock Rouge: A package for automatic evaluation of summaries.
\newblock In \emph{Text summarization branches out}, pages 74--81, 2004.

\bibitem[Lin et~al.(2014)Lin, Maire, Belongie, Hays, Perona, Ramanan, Doll{\'a}r, and Zitnick]{coco}
Tsung-Yi Lin, Michael Maire, Serge Belongie, James Hays, Pietro Perona, Deva Ramanan, Piotr Doll{\'a}r, and C~Lawrence Zitnick.
\newblock Microsoft coco: Common objects in context.
\newblock In \emph{Computer Vision--ECCV 2014: 13th European Conference, Zurich, Switzerland, September 6-12, 2014, Proceedings, Part V 13}, pages 740--755. Springer, 2014.

\bibitem[Liu et~al.(2022)Liu, Kan, Shan, and Xilin]{fcfd}
Dongyang Liu, Meina Kan, Shiguang Shan, and CHEN Xilin.
\newblock Function-consistent feature distillation.
\newblock In \emph{The Eleventh International Conference on Learning Representations}, 2022.

\bibitem[Liu et~al.(2021)Liu, Lin, Cao, Hu, Wei, Zhang, Lin, and Guo]{swin}
Ze Liu, Yutong Lin, Yue Cao, Han Hu, Yixuan Wei, Zheng Zhang, Stephen Lin, and Baining Guo.
\newblock Swin transformer: Hierarchical vision transformer using shifted windows.
\newblock In \emph{Proceedings of the IEEE/CVF international conference on computer vision}, pages 10012--10022, 2021.

\bibitem[Lu et~al.(2021)Lu, Ghaddar, Rashid, Rezagholizadeh, Ghodsi, and Langlais]{RWKD}
Peng Lu, Abbas Ghaddar, Ahmad Rashid, Mehdi Rezagholizadeh, Ali Ghodsi, and Philippe Langlais.
\newblock {RW}-{KD}: Sample-wise loss terms re-weighting for knowledge distillation.
\newblock In \emph{Findings of the Association for Computational Linguistics: EMNLP 2021}, pages 3145--3152, Punta Cana, Dominican Republic, 2021. Association for Computational Linguistics.

\bibitem[Ma et~al.(2018)Ma, Zhang, Zheng, and Sun]{shufflenet}
Ningning Ma, Xiangyu Zhang, Hai-Tao Zheng, and Jian Sun.
\newblock Shufflenet v2: Practical guidelines for efficient cnn architecture design.
\newblock In \emph{Proceedings of the European conference on computer vision (ECCV)}, pages 116--131, 2018.

\bibitem[Mirzadeh et~al.(2020)Mirzadeh, Farajtabar, Li, Levine, Matsukawa, and Ghasemzadeh]{TAKD}
Seyed~Iman Mirzadeh, Mehrdad Farajtabar, Ang Li, Nir Levine, Akihiro Matsukawa, and Hassan Ghasemzadeh.
\newblock Improved knowledge distillation via teacher assistant.
\newblock In \emph{Proceedings of the AAAI conference on artificial intelligence}, pages 5191--5198, 2020.

\bibitem[Oquab et~al.(2023)Oquab, Darcet, Moutakanni, Vo, Szafraniec, Khalidov, Fernandez, Haziza, Massa, El-Nouby, et~al.]{dinov2}
Maxime Oquab, Timoth{\'e}e Darcet, Th{\'e}o Moutakanni, Huy Vo, Marc Szafraniec, Vasil Khalidov, Pierre Fernandez, Daniel Haziza, Francisco Massa, Alaaeldin El-Nouby, et~al.
\newblock Dinov2: Learning robust visual features without supervision.
\newblock \emph{arXiv preprint arXiv:2304.07193}, 2023.

\bibitem[Radford et~al.(2019)Radford, Wu, Child, Luan, Amodei, and Sutskever]{gpt2}
Alec Radford, Jeff Wu, Rewon Child, David Luan, Dario Amodei, and Ilya Sutskever.
\newblock Language models are unsupervised multitask learners.
\newblock \emph{OpenAI blog}, 2019.

\bibitem[Romero et~al.(2014)Romero, Ballas, Kahou, Chassang, Gatta, and Bengio]{fitnet}
Adriana Romero, Nicolas Ballas, Samira~Ebrahimi Kahou, Antoine Chassang, Carlo Gatta, and Yoshua Bengio.
\newblock Fitnets: Hints for thin deep nets.
\newblock \emph{arXiv preprint arXiv:1412.6550}, 2014.

\bibitem[Sandler et~al.(2018)Sandler, Howard, Zhu, Zhmoginov, and Chen]{mobilenet}
Mark Sandler, Andrew Howard, Menglong Zhu, Andrey Zhmoginov, and Liang-Chieh Chen.
\newblock Mobilenetv2: Inverted residuals and linear bottlenecks.
\newblock In \emph{Proceedings of the IEEE conference on computer vision and pattern recognition}, pages 4510--4520, 2018.

\bibitem[Sanh et~al.(2019)Sanh, Debut, Chaumond, and Wolf]{distilbert}
Victor Sanh, Lysandre Debut, Julien Chaumond, and Thomas Wolf.
\newblock Distilbert, a distilled version of bert: smaller, faster, cheaper and lighter.
\newblock \emph{arXiv preprint arXiv:1910.01108}, 2019.

\bibitem[Shannon(1948)]{entropy}
Claude~Elwood Shannon.
\newblock A mathematical theory of communication.
\newblock \emph{The Bell system technical journal}, 27\penalty0 (3):\penalty0 379--423, 1948.

\bibitem[Simonyan and Zisserman(2015)]{vgg}
K Simonyan and A Zisserman.
\newblock Very deep convolutional networks for large-scale image recognition.
\newblock In \emph{3rd International Conference on Learning Representations (ICLR 2015)}. Computational and Biological Learning Society, 2015.

\bibitem[Sun et~al.(2024)Sun, Ren, Li, Wang, and Cao]{LS}
Shangquan Sun, Wenqi Ren, Jingzhi Li, Rui Wang, and Xiaochun Cao.
\newblock Logit standardization in knowledge distillation.
\newblock In \emph{Proceedings of the IEEE/CVF Conference on Computer Vision and Pattern Recognition}, pages 15731--15740, 2024.

\bibitem[Tian et~al.(2019)Tian, Krishnan, and Isola]{crd}
Yonglong Tian, Dilip Krishnan, and Phillip Isola.
\newblock Contrastive representation distillation.
\newblock \emph{arXiv preprint arXiv:1910.10699}, 2019.

\bibitem[Touvron et~al.(2021)Touvron, Cord, Douze, Massa, Sablayrolles, and J{\'e}gou]{deit}
Hugo Touvron, Matthieu Cord, Matthijs Douze, Francisco Massa, Alexandre Sablayrolles, and Herv{\'e} J{\'e}gou.
\newblock Training data-efficient image transformers \& distillation through attention.
\newblock In \emph{International conference on machine learning}, pages 10347--10357. PMLR, 2021.

\bibitem[Van~der Maaten and Hinton(2008)]{tsne}
Laurens Van~der Maaten and Geoffrey Hinton.
\newblock Visualizing data using t-sne.
\newblock \emph{Journal of machine learning research}, 9\penalty0 (11), 2008.

\bibitem[Wang et~al.(2022{\natexlab{a}})Wang, Kordi, Mishra, Liu, Smith, Khashabi, and Hajishirzi]{self-inst}
Yizhong Wang, Yeganeh Kordi, Swaroop Mishra, Alisa Liu, Noah~A Smith, Daniel Khashabi, and Hannaneh Hajishirzi.
\newblock Self-instruct: Aligning language models with self-generated instructions.
\newblock \emph{arXiv preprint arXiv:2212.10560}, 2022{\natexlab{a}}.

\bibitem[Wang et~al.(2022{\natexlab{b}})Wang, Mishra, Alipoormolabashi, Kordi, Mirzaei, Arunkumar, Ashok, Dhanasekaran, Naik, Stap, et~al.]{s-ni}
Yizhong Wang, Swaroop Mishra, Pegah Alipoormolabashi, Yeganeh Kordi, Amirreza Mirzaei, Anjana Arunkumar, Arjun Ashok, Arut~Selvan Dhanasekaran, Atharva Naik, David Stap, et~al.
\newblock Super-naturalinstructions: Generalization via declarative instructions on 1600+ nlp tasks.
\newblock \emph{arXiv preprint arXiv:2204.07705}, 2022{\natexlab{b}}.

\bibitem[Zagoruyko and Komodakis(2016)]{wideresnet}
Sergey Zagoruyko and Nikos Komodakis.
\newblock Wide residual networks.
\newblock In \emph{Procedings of the British Machine Vision Conference 2016}. British Machine Vision Association, 2016.

\bibitem[Zhang et~al.(2024)Zhang, Zhang, Sun, Chen, and Xu]{dskd}
Songming Zhang, Xue Zhang, Zengkui Sun, Yufeng Chen, and Jinan Xu.
\newblock Dual-space knowledge distillation for large language models.
\newblock \emph{arXiv preprint arXiv:2406.17328}, 2024.

\bibitem[Zhang et~al.(2020)Zhang, Lan, Dai, Zeng, Bai, Chang, and Wei]{PAD}
Youcai Zhang, Zhonghao Lan, Yuchen Dai, Fangao Zeng, Yan Bai, Jie Chang, and Yichen Wei.
\newblock Prime-aware adaptive distillation.
\newblock In \emph{Computer Vision--ECCV 2020: 16th European Conference, Glasgow, UK, August 23--28, 2020, Proceedings, Part XIX 16}, pages 658--674. Springer, 2020.

\bibitem[Zhao et~al.(2022)Zhao, Cui, Song, Qiu, and Liang]{dkd}
Borui Zhao, Quan Cui, Renjie Song, Yiyu Qiu, and Jiajun Liang.
\newblock Decoupled knowledge distillation.
\newblock In \emph{Proceedings of the IEEE/CVF Conference on computer vision and pattern recognition}, pages 11953--11962, 2022.

\bibitem[Zheng and Yang(2024)]{ttm}
Kaixiang Zheng and En-Hui Yang.
\newblock Knowledge distillation based on transformed teacher matching.
\newblock \emph{arXiv preprint arXiv:2402.11148}, 2024.

\bibitem[Zhu et~al.(2024)Zhu, Shang, Yuan, Zhang, Li, Li, and Jiao]{dynamickd}
Songling Zhu, Ronghua Shang, Bo Yuan, Weitong Zhang, Wenjie Li, Yangyang Li, and Licheng Jiao.
\newblock Dynamickd: An effective knowledge distillation via dynamic entropy correction-based distillation for gap optimizing.
\newblock \emph{Pattern Recognition}, 153:\penalty0 110545, 2024.

\end{thebibliography}
}

\appendix
\clearpage
\setcounter{page}{1}
\renewcommand{\thesection}{\Alph{section}}
\renewcommand\thefigure{\Alph{section}\arabic{figure}}
\renewcommand\thetable{\Alph{section}\arabic{table}}
\setcounter{page}{1}
\setcounter{section}{0}
\setcounter{figure}{0}
\setcounter{table}{0}

\maketitlesupplementary

\section{Additional Results and Comparison}

\subsection{Comparison with TTM and WTTM}
\label{sec:ttm-appendix} 
TTM \cite{ttm} removes student temperature in KD, revealing an inherent Rényi entropy regularizer that could improve generalization. WTTM further up-weights uncertain samples using the power sum of the teacher’s output, akin to our $w_{\text{base}}$\footnote{In fact, they also noted teacher entropy as a potential weighting factor, but left its systematic exploration for future work.}. Notably, TTM's Rényi entropy is an \textit{inherent effect} of its structural modification, while EA-KD \textit{actively} uses Shannon entropy to prioritize valuable samples. Moreover, unlike WTTM’s static teacher-based weighting, EA-KD dynamically adapts to the student’s learning (via $H^\mathcal{S}$), yielding stronger enhancement to TTM (\cref{tab:ea-ttm}). Finally, while their methods combine with feature-based KDs by adding and balancing their losses, EA-KD offers direct integration to both logit- and feature-based KDs through reweighting.

\begin{table}[h]
\centering
\scriptsize
\caption{\textbf{TTM, WTTM, and EA-TTM on CIFAR-100 over 5 runs.} Unlike WTTM, EA-TTM consistently improves TTM, highlighting the superiority of dynamic over static weighting.}
\setlength{\tabcolsep}{5pt} 
\label{tab:ea-ttm}
\begin{tabularx}{1.0\linewidth}{ccYYc}
\toprule
Teacher & Student & TTM & WTTM & EA-TTM (Ours) \\
\midrule
ResNet32$\times$4 & ResNet8$\times$4& \ul{76.17}$\pm$0.28 & 76.06$\pm$0.27 & \textbf{76.25}$\pm$0.04 \\
WRN-40-2 & WRN-40-1 & 74.32$\pm$0.31 & \ul{74.58}$\pm$0.26 & \textbf{74.61}$\pm$0.07\\
ResNet32$\times$4 & SN-V2 & \ul{76.57}$\pm$0.26 &  76.55$\pm$0.08 & \textbf{76.65}$\pm$0.16 \\
\bottomrule
\end{tabularx}
\end{table}

\subsection{MS-COCO} \cref{tab:cocotable2} reports additional results for MS-COCO, where EA-DKD is compared against DKD \cite{dkd} using a ResNet-101 to ResNet-18 distillation setting. The results show that EA-DKD consistently improves upon DKD across all evaluation metrics.

\begin{table}[h]
\scriptsize
\centering
\caption{\textbf{More Results on MS-COCO.}}
\begin{tabularx}{1.0\linewidth}{YYYYY}
\toprule
R101$\to$R18 & AP & AP\(_{50}\) & AP\(_{75}\) \\
\midrule
DKD [39] & 35.05 & 56.60 & 37.54 \\
EA-DKD & \textbf{35.16}\color[HTML]{008F00}\texttt{+}0.11 & \textbf{56.75} \color[HTML]{008F00}\texttt{+}0.15 & \textbf{37.82} \color[HTML]{008F00}\texttt{+}0.28 \\
\bottomrule
\end{tabularx}
\label{tab:cocotable2}
\end{table}

\subsection{LLM Distillation}
\cref{tab:mllm} extends the comparison of EA-KD with Skewed KLD (SKLD) and SRKLD from the recent DistiLLM \cite{distillm} method, under the same off-policy setting without pre-training corpus. EA-KD outperforms these stronger methods on most datasets, highlighting its overall effectiveness.

\begin{table}[h]
\centering
\scriptsize
\setlength{\tabcolsep}{3pt}
\caption{{\textbf{More LLM Distillation Results.}}}
\label{tab:mllm}
\begin{tabularx}{\linewidth}{cYYYYYY}
\toprule
Method & Dolly & SInst & Vicuna & S-NI & UnNI & Avg. \\
\midrule
SKLD (DistiLLM) & 24.03 & \textbf{10.66} & 14.70 & \underline{17.99} & \underline{21.18} & \underline{17.71} \\
SRKLD (DistiLLM) & \underline{24.48} & \underline{10.35} & \underline{14.88} & 16.53 & 19.68 & 17.19 \\
\midrule
EA-KD & \textbf{24.95} & 10.59 & \textbf{16.41} & \textbf{18.27} & \textbf{21.46} & \textbf{18.34} \\
\bottomrule
\end{tabularx}
\end{table}

\section{Additional Analysis and Visualizations} 

This section provides additional analysis to complement the main paper. It includes a detailed examination of high student entropy samples, loss landscape comparisons for KD and EA-KD, and t-SNE visualizations for various KD frameworks and their EA-enhanced variants.

\subsection{Analysis of High Student Entropy Samples} 
\label{app:high_Hs}
\begin{figure*}
\centering
\subcaptionbox{Accuracy vs. Student Entropy Segments. \label{fig:student_entropy_accuracy}}{
    \includegraphics[height=11.5\baselineskip]{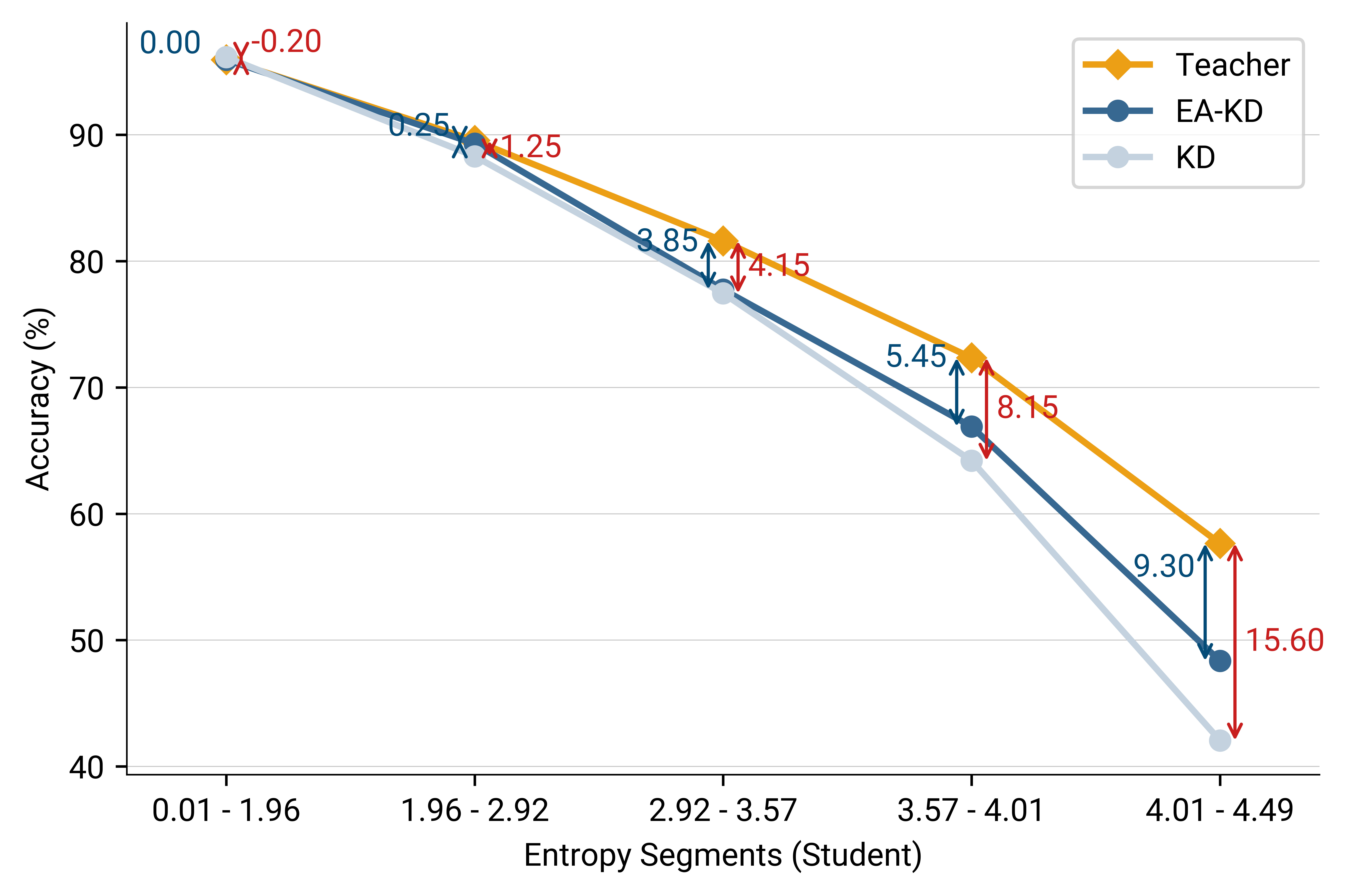}
}
\hfill
\subcaptionbox{t-SNE of Students Across Training Epochs.\label{fig:tsne_epoch}}{
    \includegraphics[height=11.5\baselineskip]{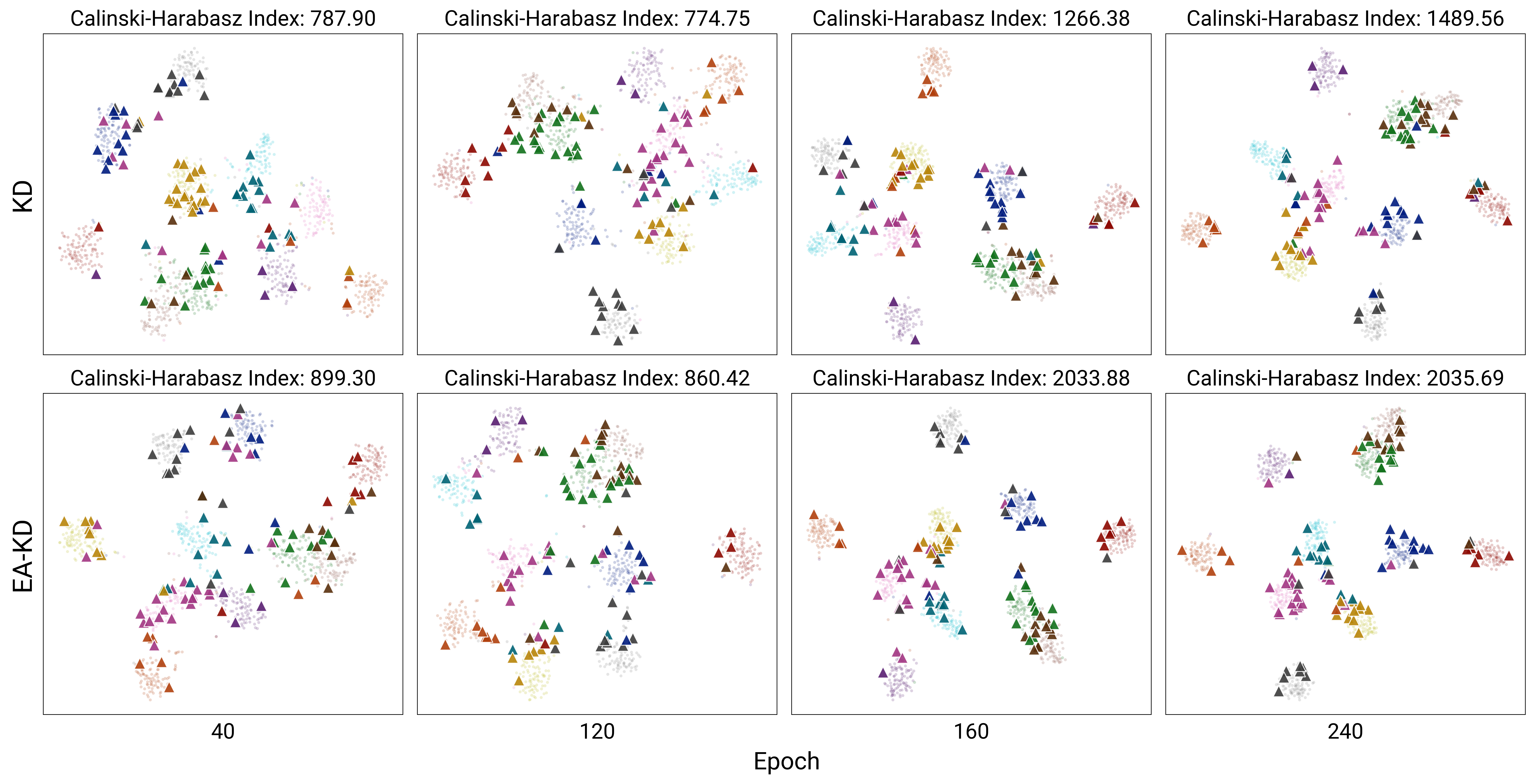}
}
\caption{
\textbf{High Student Entropy Analysis in KD and EA-KD.} (a) Higher student entropy samples correlate with larger accuracy gaps in KD. EA-KD demonstrates closer alignment with the teacher, particularly in high-entropy regions. (b) High $H^\mathcal{S}$ samples consistently cluster near decision boundaries, reflecting the student’s real-time learning needs. EA-KD adapts to this dynamic, achieving enhanced class separability over epochs.
}
\label{fig:student_entropy_combined}
\end{figure*}

Complementary to the teacher entropy analysis in \cref{sec:intro}, this section emphasizes the critical role of high student entropy samples in the adaptive reweighting process of EA-KD. As shown in \cref{fig:student_entropy_accuracy}, these samples correlate with larger teacher-student accuracy gaps similar to the teacher entropy segments (\cref{fig:acc_vs_entropy}). Additionally, \cref{fig:tsne_epoch} presents the students' t-SNE visualizations across training epochs for KD and EA-KD, with high $H^\mathcal{S}$ and high $w_{\text{EA}}$ samples highlighted, respectively. Notably, these samples also cluster near decision boundaries similar to the teacher’s t-SNE (\cref{fig:tsne_t}). Unlike the teacher, however, the student’s top-entropy samples shift dynamically over training. For instance, in KD, the yellow class at epoch 40 is located near the center of most clusters and contains numerous high-entropy samples. By epoch 120, high-entropy samples are more prevalent in the green and purple classes as they shift closer to the center. With this dynamic integrated into $w_{\text{EA}}$, EA-KD continuously adapts to the student’s evolving learning needs throughout training, complementing the static nature of $w_{\text{base}}$ and fostering a tailored knowledge transfer.

\begin{figure*}
    \centering
    \includegraphics[width=1.0\textwidth]{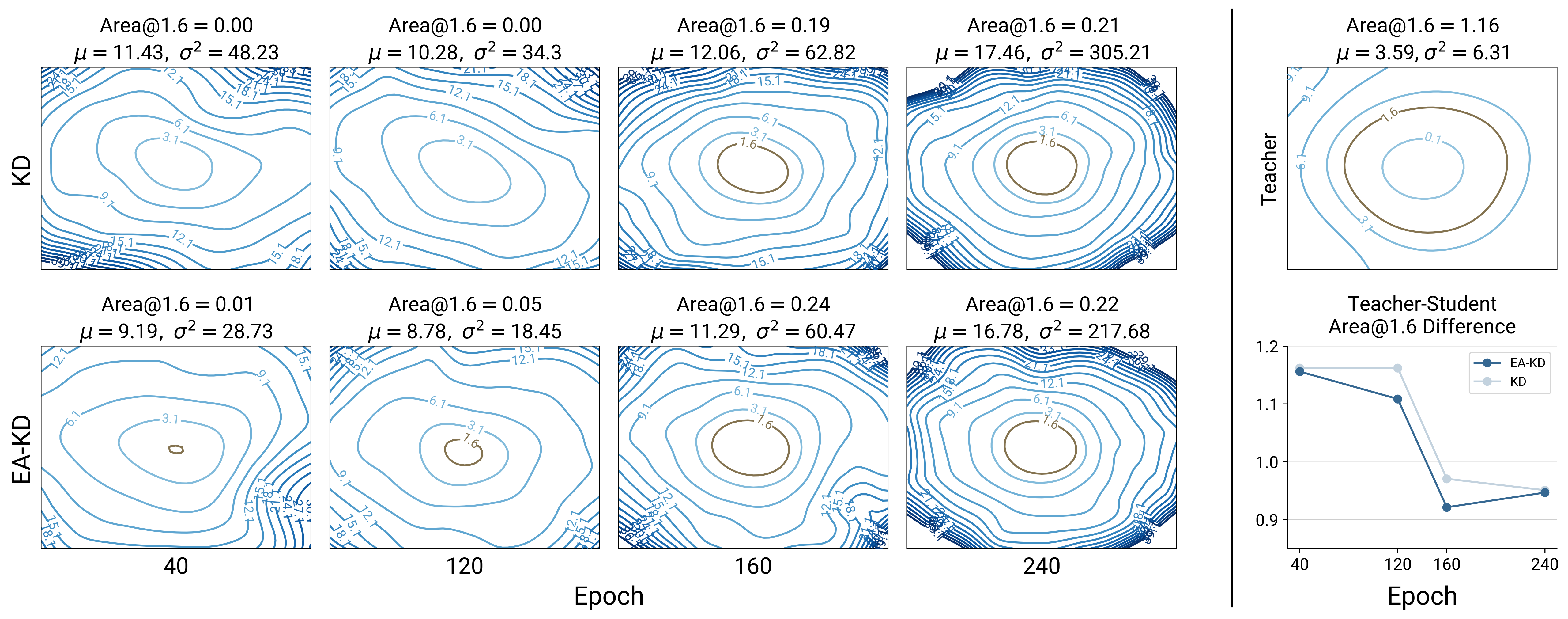}
    \caption{\textbf{Loss Surface and Differences in Area@1.6 for KD and EA-KD Students Across Epochs.} Contour plots illustrate the loss landscapes of KD (first row) and EA-KD (second row) across training epochs. The line plot (lower right) tracks the differences in Area@1.6 between the teacher and students over epochs. EA-KD exhibits a more stable and robust loss surface, with greater Area@1.6 earlier in training, signifying a more efficient learning process.}
    \label{fig:kd-landscapes}
\end{figure*}

\subsection{Loss Landscape Analysis for KD and EA-KD}
We further analyze the robustness of KD and EA-KD by examining their loss landscapes across training epochs. As shown in \cref{fig:kd-landscapes}, EA-KD achieves a consistently smoother loss surface compared to KD (\eg $\sigma^2$ of 217.68 vs. 305.21 at epoch 240) and larger Area@1.6 values. Notably, KD’s Area@1.6 metric remains 0 at epochs 40 and 120, indicating the absence of low-loss regions. In contrast, EA-KD achieves values of 0.01 and 0.05, respectively, underscoring its improved learning efficiency by enabling the student to assimilate the key knowledge. These results highlight EA-KD’s ability to ensure a more efficient optimization, facilitating enhanced generalizability compared to standard KD.
\vfill
\subsection{t-SNE of KD Frameworks and EA-methods}
\begin{figure*}[t]
    \centering
    \includegraphics[width=.75\textwidth]{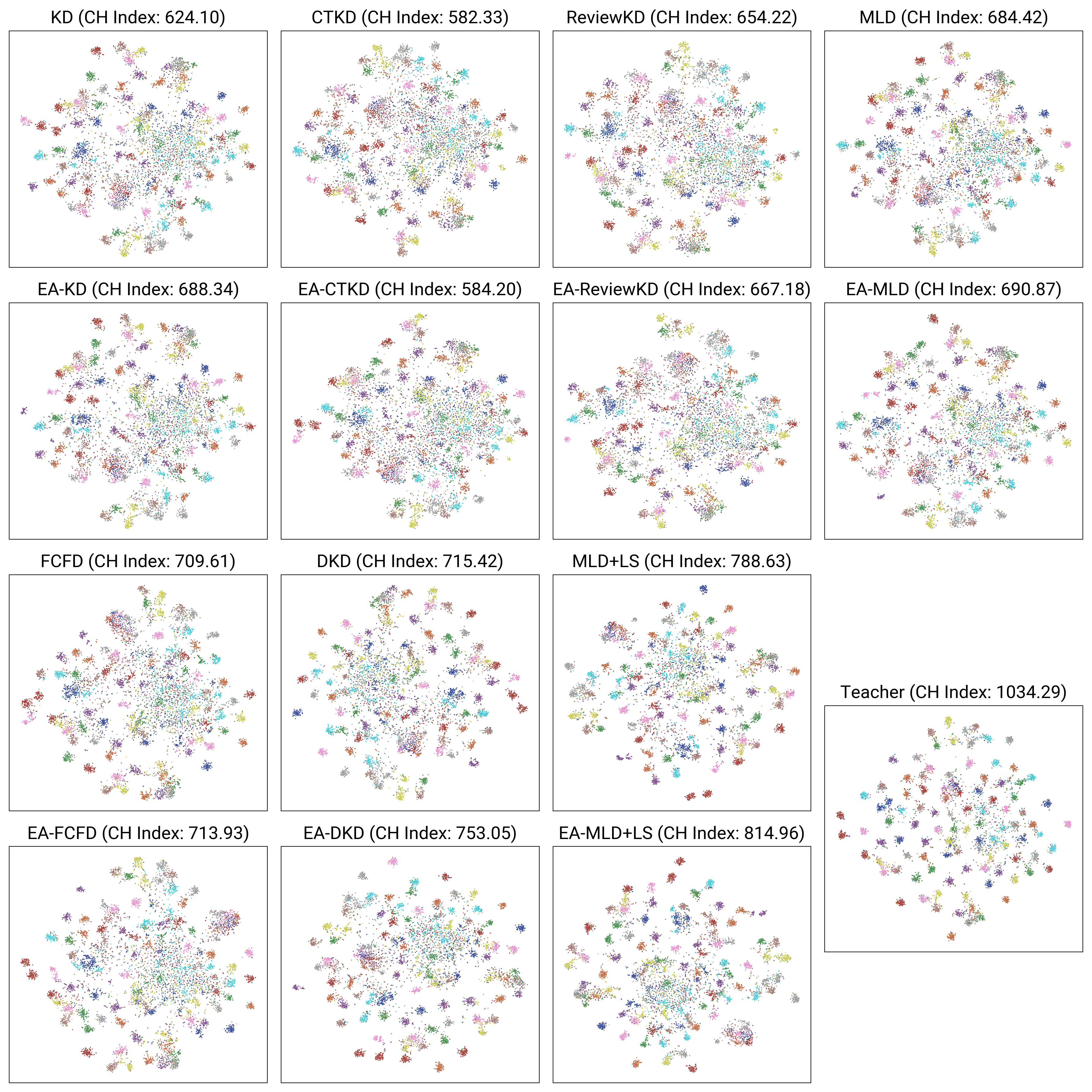}
    \caption{\textbf{t-SNE Visualizations of EA-methods vs. Baselines.} The students from various KD frameworks (first and third row) and their EA-enhanced counterparts (second and fourth row), along with the teacher (lower right), are shown. EA-methods consistently achieve higher CH indices, indicating better class separability.}
    \label{fig:tsne_baseline}
\end{figure*}

The t-SNE visualizations of students from various KD frameworks and their EA-method variants, alongside the teacher’s representation, are shown in \cref{fig:tsne_baseline}. EA-methods consistently exhibit more distinct and well-defined class clusters compared to their respective baselines, as evidenced by higher CH indices. Remarkably, the SOTA EA-MLD+LS achieves the highest CH index of 814.96, indicating its superior performance and closer alignment with the teacher. Furthermore, although EA-DKD ranks fourth in performance (\cref{tab:cifar100}), it demonstrates the second-best class separability, underscoring its enhanced robustness as discussed in \cref{sec:analysis}. These findings highlight EA-KD’s versatility in enhancing class separability and improving knowledge transfer across diverse KD frameworks.

\vfill
\section{Statistics for Table 2 and 4}
\cref{tab:cifar100std} and \cref{tab:imagenetstd} present the mean and standard deviation of EA-methods on CIFAR-100 and ImageNet across multiple runs, respectively. Since baseline papers typically report only mean values, formal statistical tests such as p-value computation are not applicable. Nonetheless, EA-methods consistently demonstrate improved performance with low variance, highlighting the stable and reliable gains.

\begin{table*}[ht]
\centering
\footnotesize  
\setlength{\tabcolsep}{3pt} 

\begin{threeparttable}
\caption{\textbf{Results on CIFAR-100.} Mean accuracy (\%) and standard deviation across five runs are reported.}
\label{tab:cifar100std}
\begin{tabularx}{\textwidth}{clYYYcYYY@{}}
\toprule
\multicolumn{1}{c}{\multirow{4}{*}{Type}} & \multicolumn{1}{c}{\multirow{2}{*}{Teacher}} & ResNet32$\times$4 & WRN-28-4 & WRN-40-2 & VGG13 & VGG13 & ResNet50 & ResNet32$\times$4 \\
& & 79.42 & 78.60 & 75.61 & 74.64 & 74.64 & 79.34 & 79.42 \\
& \multicolumn{1}{c}{\multirow{2}{*}{Student}} & ResNet8$\times$4 & WRN-16-2 & WRN-40-1 & VGG8 & MN-V2 & MN-V2 & SN-V2 \\
& & 72.50 & 73.26 & 71.98 & 70.36 & 64.60 & 64.60 & 71.82 \\ 
\midrule

\mrn{4}{2}{Logit} & EA-KD & 75.46$\pm$0.15 & 75.79$\pm$0.14 & 74.38$\pm$0.10 & 74.08$\pm$0.10 & 69.17$\pm$0.08 & 69.67$\pm$0.26 & 75.91$\pm$0.25 \\

& EA-CTKD & 75.18$\pm$0.24 & 75.72$\pm$0.16 & 74.03$\pm$0.05 & 73.79$\pm$0.10 & 69.19$\pm$0.26 & 69.38$\pm$0.36 & 76.02$\pm$0.15 \\

& EA-DKD & 76.80$\pm$0.05 & 76.74$\pm$0.09 & 74.98$\pm$0.15 & 75.07$\pm$0.14 & 70.39$\pm$0.14 & 70.98$\pm$0.13 & 77.72$\pm$0.07 \\

& EA-MLD & 77.65$\pm$0.05 & \underline{77.47}$\pm$0.12 & \underline{75.77}$\pm$0.21 & 75.28$\pm$0.22 & 70.72$\pm$0.15 & \underline{71.43}$\pm$0.18 & \underline{78.85}$\pm$0.05 \\

& EA-MLD+LS & \textbf{78.38}$\pm$0.10 & \textbf{77.60}$\pm$0.07 & \textbf{75.78}$\pm$0.10 & \textbf{75.38}$\pm$0.15 & \ul{70.67}$\pm$0.25 & 71.36$\pm$0.21 & \textbf{79.13}$\pm$0.23 \\

\midrule

\mr{2}{Feature} 
& EA-ReviewKD & 76.10$\pm$0.13 & 76.95$\pm$0.16 & 75.43$\pm$0.13 & 74.56$\pm$0.11 & 70.55$\pm$0.09 & 69.80$\pm$0.18 & 78.22$\pm$0.15 \\

& EA-FCFD & 77.50$\pm$0.08 & 77.15$\pm$0.15 & 75.30$\pm$0.03 & \underline{75.36}$\pm$0.06 & \textbf{71.02}$\pm$0.26 & \textbf{71.97}$\pm$0.29 & 78.75$\pm$0.32 \\

\bottomrule
\end{tabularx}
\end{threeparttable}
\end{table*} 

\begin{table*}[t]
\footnotesize
\centering
\caption{\textbf{Results on ImageNet.} Mean accuracy (\%) and standard deviation across three runs are reported for EA-methods.}
\begin{tabularx}{1\textwidth}{ccYcYYccccc}
\toprule
Teacher & Student & KD [10] & EA-KD & KD+LS [30] & DKD [39] & EA-DKD & DKD+LS [30] & EA-DKD+LS & PAD [38]  \\
\cmidrule(){1-2} \cmidrule(lr){3-5} \cmidrule(lr){6-9} \cmidrule(lr){10-10} \cmidrule(){11-11}
73.31 & 69.75 & 71.03 & \textbf{71.79}$\pm$0.02  & \ul{71.42} & 71.70 & \ul{71.96}$\pm$0.08  & 71.88 & \textbf{71.99}$\pm$0.01  & 71.71  \\
\bottomrule
\end{tabularx}
\label{tab:imagenetstd}
\end{table*}

\end{document}